\documentclass[11pt]{report}

\RequirePackage[authoryear,round]{natbib}

\RequirePackage[hyphens]{url}
\usepackage[pdftex]{hyperref}
\hypersetup{
    unicode=false,
    pdftoolbar=true,
    pdfmenubar=true,
    pdffitwindow=false,
    pdfnewwindow=true,
    colorlinks=true,
    linkcolor=MyGrey,
    citecolor=CitationColor,
    filecolor=DarkGreen,
    urlcolor=CitationColor,
    pdftitle={},
    pdfauthor={},
}

\usepackage[skip=2mm,indent=5mm]{parskip}
\usepackage[small]{caption}
\RequirePackage{amsthm,amsmath,amsfonts,amssymb}

\newenvironment{myquote}{%
   \list{}{\rightmargin\leftmargin}\item\relax
}{\endlist}

\newcommand{\paperIcitation}{\textbf{Baumann}\textbf{, J.}, \& Loi, M. (2023). Fairness and risk: an ethical argument for a group fairness definition insurers can use. \textit{Philosophy \& Technology, 36}(3), 45.}
\newcommand{\paperIIcitation}{\textbf{Baumann}\footnote{\label{equalcont}Equal contribution.}\textbf{, J.}, Hertweck\textsuperscript{\ref{equalcont}}, C., Loi, M., \& Heitz, C. (2023). Unification, Extension, and Interpretation of Group Fairness Metrics for ML-Based Decision-Making. In \textit{Proceedings of the 2nd European Workshop on Algorithmic Fairness (EWAF)}.}
\newcommand{\paperIIIcitation}{\textbf{Baumann}\textbf{, J.}, Castelnovo, A., Crupi, R., Inverardi, N., \& Regoli, D. (2023). Bias on demand: a modelling framework that generates synthetic data with bias. In \textit{Proceedings of the 2023 ACM Conference on Fairness, Accountability, and Transparency (FAccT)} (pp. 1002-1013).}
\newcommand{\paperIVcitation}{Pagan\textsuperscript{\ref{equalcont}}, N., \textbf{Baumann}\textsuperscript{\ref{equalcont}}\textbf{, J.}, Elokda, E., De Pasquale, G., Bolognani, S., \& Hannák, A. (2023). A classification of feedback loops and their relation to biases in automated decision-making systems. In \textit{Proceedings of the 3rd ACM Conference on Equity and Access in Algorithms, Mechanisms, and Optimization (EAAMO)} (pp. 1-14).}
\newcommand{\paperVcitation}{\textbf{Baumann}\textbf{, J.}, Sapiezynski, P., Heitz, C., \& Hannák, A. (2024). Fairness in online ad delivery. In \textit{Proceedings of the 2024 ACM Conference on Fairness, Accountability, and Transparency (FAccT)} (pp. 1418-1432).}
\newcommand{\paperVIcitation}{\textbf{Baumann}\textbf{, J.}, Hannák, A., \& Heitz, C. (2022). Enforcing group fairness in algorithmic decision making: Utility maximization under sufficiency. In \textit{Proceedings of the 2022 ACM Conference on Fairness, Accountability, and Transparency (FAccT)} (pp. 2315-2326).}
\newcommand{\paperVIIcitation}{\textbf{Baumann}\textbf{, J.}, \& Mendler-Dünner, C. (2024). Algorithmic Collective Action in Recommender Systems: Promoting Songs by Reordering Playlists. In \textit{The Thirty-eighth Annual Conference on Neural Information Processing Systems (NeurIPS)}.}
\newcommand{\paperVIIIcitation}{Vajiac\textsuperscript{\ref{equalcont}}, C., Frey\textsuperscript{\ref{equalcont}}, A., \textbf{Baumann}\textsuperscript{\ref{equalcont}}\textbf{, J.}, Smith\textsuperscript{\ref{equalcont}}, A., Amarasinghe, K., Lai, A., Rodolfa, K.T.,  \& Ghani, R. (2024). Preventing eviction-caused homelessness through ML-informed distribution of rental assistance. In \textit{Proceedings of the AAAI Conference on Artificial Intelligence (AAAI)} (Vol. 38, No. 20, pp. 22393-22400).}

\newcommand{\paperIcitationnew}{\textbf{Baumann}\textbf{, J.}, \& Loi, M. (2023). Fairness and risk: an ethical argument for a group fairness definition insurers can use. \textit{Philosophy \& Technology, 36}(3), 45. URL \url{https://doi.org/10.1007/s13347-023-00624-9}}
\newcommand{\paperIIcitationnew}{\textbf{Baumann}\footnote{\label{equalcontributionI}Equal contribution.}\textbf{, J.}, Hertweck\textsuperscript{\ref{equalcontributionI}}, C., Loi, M., \& Heitz, C. (2023). Unification, Extension, and Interpretation of Group Fairness Metrics for ML-Based Decision-Making. In \textit{Proceedings of the 2nd European Workshop on Algorithmic Fairness (EWAF)}. URL \url{https://ceur-ws.org/Vol-3442/paper-23.pdf}}
\newcommand{\paperIIIcitationnew}{\textbf{Baumann}\textbf{, J.}, Castelnovo, A., Crupi, R., Inverardi, N., \& Regoli, D. (2023). Bias on demand: a modelling framework that generates synthetic data with bias. In \textit{Proceedings of the 2023 ACM Conference on Fairness, Accountability, and Transparency (FAccT)} (pp. 1002-1013). URL \url{https://doi.org/10.1145/3593013.3594058}}
\newcommand{\paperIVcitationnew}{Pagan\footnote{\label{equalcontributionII}Equal contribution.}, N., \textbf{Baumann}\textsuperscript{\ref{equalcontributionII}}\textbf{, J.}, Elokda, E., De Pasquale, G., Bolognani, S., \& Hannák, A. (2023). A classification of feedback loops and their relation to biases in automated decision-making systems. In \textit{Proceedings of the 3rd ACM Conference on Equity and Access in Algorithms, Mechanisms, and Optimization (EAAMO)} (pp. 1-14). URL \url{https://doi.org/10.1145/3617694.3623227}}
\newcommand{\paperVcitationnew}{\textbf{Baumann}\textbf{, J.}, Sapiezynski, P., Heitz, C., \& Hannák, A. (2024). Fairness in online ad delivery. In \textit{Proceedings of the 2024 ACM Conference on Fairness, Accountability, and Transparency (FAccT)} (pp. 1418-1432). URL \url{https://doi.org/10.1145/3630106.3658980}}
\newcommand{\paperVIcitationnew}{\textbf{Baumann}\textbf{, J.}, Hannák, A., \& Heitz, C. (2022). Enforcing group fairness in algorithmic decision making: Utility maximization under sufficiency. In \textit{Proceedings of the 2022 ACM Conference on Fairness, Accountability, and Transparency (FAccT)} (pp. 2315-2326). URL \url{https://doi.org/10.1145/3531146.3534645}}
\newcommand{\paperVIIcitationnew}{\textbf{Baumann}\textbf{, J.}, \& Mendler-Dünner, C. (2024). Algorithmic Collective Action in Recommender Systems: Promoting Songs by Reordering Playlists. In \textit{The Thirty-eighth Annual Conference on Neural Information Processing Systems (NeurIPS)}. URL \url{https://proceedings.neurips.cc/paper_files/paper/2024/file/d79792543133425ff79513c147dc8881-Paper-Conference.pdf}}
\newcommand{\paperVIIIcitationnew}{Vajiac\footnote{\label{equalcontributionIII}Equal contribution.}, C., Frey\textsuperscript{\ref{equalcontributionIII}}, A., \textbf{Baumann}\textsuperscript{\ref{equalcontributionIII}}\textbf{, J.}, Smith\textsuperscript{\ref{equalcontributionIII}}, A., Amarasinghe, K., Lai, A., Rodolfa, K.T.,  \& Ghani, R. (2024). Preventing eviction-caused homelessness through ML-informed distribution of rental assistance. In \textit{Proceedings of the AAAI Conference on Artificial Intelligence (AAAI)} (Vol. 38, No. 20, pp. 22393-22400). URL \url{https://ojs.aaai.org/index.php/AAAI/article/view/30246}}

\RequirePackage{graphicx}
\usepackage{subcaption}
\usepackage{comment}
\usepackage{multicol}
\usepackage[svgnames]{xcolor}
\usepackage{pdfpages}

\usepackage[dvipsnames]{xcolor}
\definecolor{MyGrey}{rgb}{0,0.06,0.21}
\definecolor{MyYellow}{rgb}{0.51,0.47,0.21}
\color{MyGrey}

\usepackage{fullpage}
\usepackage[utf8]{inputenc}
\usepackage[T1]{fontenc}
\usepackage{microtype}
\usepackage{csquotes}
\usepackage[osf,sc]{mathpazo}
\RequirePackage[scaled=0.90]{helvet}
\RequirePackage[scaled=0.85]{beramono}
\RequirePackage{textcomp}

\usepackage{tikz}
\usepackage{subcaption}
\usepackage{enumitem}

\usepackage{booktabs}

\usepackage{tcolorbox}

\usepackage[symbol]{footmisc}

\definecolor{CitationColor}{RGB}{115, 78, 0}
\definecolor{DarkGreen}{RGB}{98,179,149}
\definecolor{DarkRed}{RGB}{206,86,69}

\definecolor{goalIcolor}{RGB}{4,146,122}

\definecolor{goalIIcolor}{RGB}{30,136,229}

\definecolor{goalIIIcolor}{RGB}{216,27,96}

\usepackage{titlesec}

\titleformat{\chapter}[display]
    {\normalfont\LARGE}{Chapter \thechapter}{10pt}{\Huge\bfseries}

\titlespacing*{\chapter}{0pt}{40pt}{40pt}

\titleclass{\subsubsubsection}{straight}[\subsection]

\newcounter{subsubsubsection}[subsubsection]
\renewcommand\thesubsubsubsection{\thesubsubsection.\arabic{subsubsubsection}}

\titleformat{\subsubsubsection}
  {\normalfont\normalsize\bfseries}{\thesubsubsubsection}{1em}{}
\titlespacing*{\subsubsubsection}
{0pt}{3.25ex plus 1ex minus .2ex}{1.5ex plus .2ex}

\makeatletter
\renewcommand\paragraph{\@startsection{paragraph}{5}{\z@}%
  {3.25ex \@plus1ex \@minus.2ex}%
  {-1em}%
  {\normalfont\normalsize\bfseries}}
\renewcommand\subparagraph{\@startsection{subparagraph}{6}{\parindent}%
  {3.25ex \@plus1ex \@minus .2ex}%
  {-1em}%
  {\normalfont\normalsize\bfseries}}
\def\toclevel@subsubsubsection{4}
\def\toclevel@paragraph{5}
\def\toclevel@subparagraph{6}
\def\l@subsubsubsection{\@dottedtocline{4}{7em}{4em}}
\def\l@paragraph{\@dottedtocline{5}{10em}{5em}}
\def\l@subparagraph{\@dottedtocline{6}{14em}{6em}}
\makeatother

\begin{document}

\begin{titlepage}
    \centering
    \includegraphics[width=0.5\textwidth]{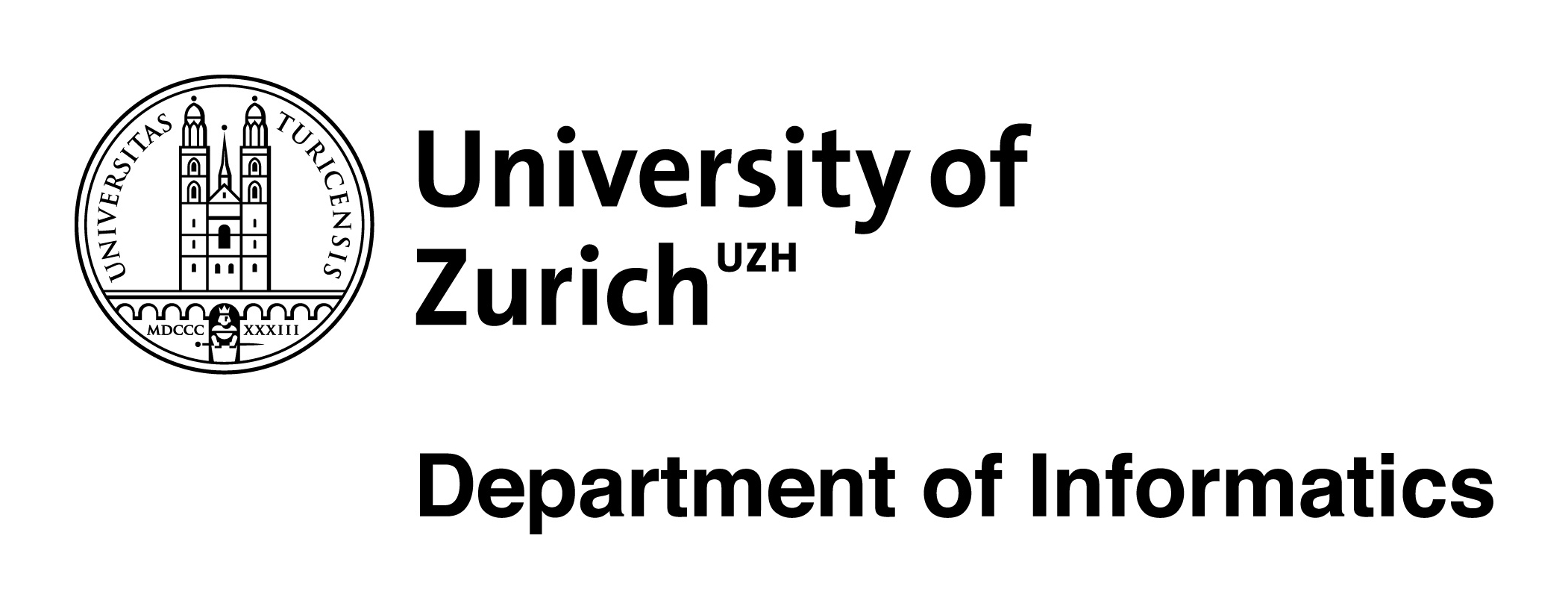}\\[1.5cm]
    
    {\LARGE\bfseries On the Societal Impact of Machine Learning\\}
    \vspace{2cm}
    
    {\large 
    Dissertation submitted to the\\
    Faculty of Business, Economics and Informatics\\
    of the University of Zurich\\[1cm]

    to obtain the degree of\\
    Doktor der Wissenschaften, Dr. sc.\\
    (corresponds to Doctor of Science, PhD)\\[1cm]
    
    presented by\\
    \textbf{Joachim Baumann}\\
    from Winterthur, ZH, Switzerland\\[1cm]
    }
    
    {\large
    approved in June 2025\\[1cm]
    
    at the request of\\
    Prof. Anik\'{o} Hann\'{a}k, PhD\\
    Prof. Mykola Pechenizkiy, PhD\\
    Prof. Celestine Mendler-Dünner, PhD\\
    Prof. Christoph Heitz, PhD\\
    }
    
\end{titlepage}

\clearpage

\pagenumbering{roman}

\noindent
The Faculty of Business, Economics and Informatics of the University of Zurich hereby authorizes the printing of this dissertation, without indicating an opinion of the views expressed in the work.
\\\\
\noindent
Zurich, July 16, 2025
\\\\
\noindent
The Chairperson of the Doctoral Board: Prof. Dr. Elaine Huang

\clearpage

\addcontentsline{toc}{section}{Abstract}
\section*{Abstract}

This thesis investigates the societal impact of machine learning (ML). ML increasingly informs consequential decisions and recommendations, significantly affecting many aspects of our lives. As these data-driven systems are typically not designed with the primary goal of having a positive societal impact, they carry the risk of discriminatory effects.~\looseness=-1

This thesis approaches the societal impact of ML through three interconnected goals. First, we establish measurement frameworks to quantify ML's impact on society in terms of the unfairness caused. Namely, we derive appropriate fairness criteria tailored to specific application domains like insurance pricing. Additionally, we connect fairness criteria to philosophical theories, revealing the normative choices that are implicit in technical implementations of fairness. Second, we decompose complex ML systems. The decomposition of ML system components enables us to systematically analyze different types of bias in isolation, shedding light on their interconnection with fairness metrics and model performance. Furthermore, we classify feedback loops in ML systems, providing a rigorous taxonomy that explains how biases are amplified over time through the dynamics between algorithms and their environment. Third, we develop practical interventions. Specifically, we derive optimal solutions for decision-making under various fairness constraints. We identify cases where supposedly fairness-enhancing system constraints make certain subgroups significantly worse off, raising questions of the appropriateness of such interventions. We propose algorithmic collective action strategies that leverage coordinated user behavior to indirectly steer recommendation systems when direct model modification is not possible. This approach can counter power imbalances and make recommendations on music streaming platforms more equitable for emerging artists. In the final part of this thesis, we demonstrate how responsibly developed ML systems can benefit society when designed with appropriate objectives. Namely, in collaboration with Allegheny County's Department of Human Services, we build a proactive outreach system that helps distribute rental assistance more effectively. We also provide ethical considerations that inform the responsible development of ML systems as decision support tools in real-world, resource-constrained contexts.~\looseness=-1

The contributions in this thesis enable more appropriate measurement of fairness in ML systems, meaningful decomposition of ML systems to anticipate bias dynamics, and introduce effective interventions that reduce algorithmic discrimination while maintaining system utility. We conclude by discussing continuous challenges and future research directions as ML systems become increasingly integrated into society, e.g., in the form of generative artificial intelligence (AI). This work offers a foundation for ensuring that ML's societal impact aligns with broader social values.~\looseness=-1

\clearpage

\addcontentsline{toc}{section}{Zusammenfassung}
\section*{Zusammenfassung}

Diese Dissertation untersucht die Auswirkungen des maschinellen Lernens (ML) auf die Gesellschaft. ML beeinflusst zunehmend Entscheidungen und Empfehlungen mit weitreichenden Folgen, und beeinflusst so viele Aspekte unseres Lebens massgeblich. Da diese datenbasierten Systeme typischerweise nicht mit dem primären Ziel entwickelt werden, positive gesellschaftliche Auswirkungen zu erzielen, bergen sie das Risiko diskriminierender Effekte.

Diese Dissertation befasst sich mit den gesellschaftlichen Auswirkungen von ML anhand von drei miteinander verbundenen Zielen. Erstens entwickeln wir ein Framework, um die Auswirkungen von ML auf die Gesellschaft in Bezug auf die dabei verursachte Unfairness zu quantifizieren. Insbesondere leiten wir Fairness-Kriterien her, die auf spezifische Anwendungsbereiche wie die Preisgestaltung von Versicherungen zugeschnitten sind. Darüber hinaus verknüpfen wir Fairness-Kriterien mit philosophischen Theorien und zeigen die normativen Entscheidungen auf, die in technischen Implementierungen von Fairness implizit enthalten sind. Zweitens werden komplexe ML-Systeme heruntergebrochen. Die Dekomposition von ML-Systemkomponenten ermöglicht es uns, verschiedene Arten von Bias (Verzerrungen) systematisch und isoliert zu analysieren und ihren Zusammenhang mit Fairness-Metriken und der Modellperformance aufzuzeigen. Darüber hinaus klassifizieren wir Feedback Loops in ML-Systemen und liefern eine fundierte Taxonomie, die erklärt, wie Bias durch die Dynamik zwischen Algorithmen und ihrer Umgebung über die Zeit verstärkt werden kann. Drittens entwickeln wir praxisorientierte Interventionen. Insbesondere leiten wir optimale Lösungen für die Entscheidungsfindung unter verschiedenen Fairness-Bedingungen ab. Wir identifizieren Fälle, in denen vermeintlich fairnessfördernde Systembedingungen bestimmte Untergruppen erheblich benachteiligen, was Fragen zur Angemessenheit solcher Interventionen aufwirft. Wir schlagen Strategien für algorithmisches kollektives Handeln vor, wobei Empfehlungssysteme indirekt mit koordiniertem Nutzerverhalten beinflusst werden, wenn eine direkte Modellmodifikation nicht möglich ist. Dieser Ansatz kann Machtungleichgewichten entgegenwirken und die Empfehlungen von Songs aufstrebender Künstlerinnen und Künstler auf Musik-Streaming-Plattformen fairer gestalten. Im letzten Teil dieser Dissertation zeigen wir, inwiefern sich verantwortungsvoll entwickelte ML-Systeme positiv auf die Gesellschaft auswirken können, sofern sie mit angemessenen Zielen konzipiert werden. In Zusammenarbeit mit dem Department of Human Services von Allegheny County entwickeln wir ein datenbasiertes System zur proaktiven Kontaktaufnahme, das dabei hilft, finanzielle Unterstützung für Mietkosten effektiver zu verteilen. Darüber hinaus präsentieren wir auch ethische Überlegungen und Erkenntnisse, die die verantwortungsvolle Entwicklung von ML-Systemen zur Entscheidungsunterstützung in realen, ressourcenbeschränkten Kontexten beeinflussen können.~\looseness=-1

Die Beiträge in dieser Dissertation ermöglichen eine adäquatere Messung von Fairness in der Nutzung von ML-Systemen, eine sinnvolle Dekomposition solcher Systeme zur Antizipation von Bias-Dynamiken und zeigen effektive Interventionen auf, die die von Algorithmen ausgehende Diskriminierung reduzieren und gleichzeitig die Nutzbarkeit des Systems erhalten. Wir konkludieren mit einer Diskussion offener Herausforderungen und zukünftiger Forschungsrichtungen, wobei wir insbesondere auf die zunehmende Integration von ML-Systemen in die Gesellschaft eingehen, z.B. in Form von generativer künstlicher Intelligenz. Diese Arbeit bietet eine fundierte Grundlage, um sicherzustellen, dass die Auswirkungen von ML im Einklang mit übergeordneten gesellschaftlichen Werten stehen.

\clearpage

\addcontentsline{toc}{section}{Acknowledgments}
\section*{Acknowledgments}

{\itshape

I would like to express my deepest gratitude to
\begin{itemize}[leftmargin=0.8cm, rightmargin=0cm, label={}]
    \setlength{\baselineskip}{0.91\baselineskip}
    \item \hspace{-5mm} Anik\'{o}, for the most incredible supervision I could have wished for, for empowering me to freely explore any research direction that captured my interest, and for all our enlightening conversations about both research and life that have been absolutely invaluable.
    \item \hspace{-5mm} Christoph, for the trust you put in me to embark on this exciting interdisciplinary research journey together and for the countless brainstorming sessions on how to build fairer machine learning models.
    \item \hspace{-5mm} Celestine, for your truly exceptional mentorship and guidance, and for generously sharing your wisdom on conducting fundamental machine learning research.
    
    \item \hspace{-5mm} the mentors I have been lucky to find:
Dirk,
for teaching me about NLP along with perfectly complementing negroni sbagliati;
Michele,
for encouraging me to embrace interdisciplinarity, and for offering me a wonderful glimpse into the fascinating world of philosophy;
Moritz,
for offering illuminating perspectives on machine learning, benchmarking, and the profound essence of being a researcher;
Mykola,
for the fruitful discussions at conferences and the insightful feedback you have provided on my work;
Piotr,
for inspiring, formative discussions and exceptional guidance throughout many Zoom meetings;
Rayid and Kit,
for sharing your invaluable expertise and hosting memorable happy hours.
    
    \item \hspace{-5mm} the SCG (%
Aleksandra,
Andrea,
Anik\'{o},
Azza,
Corinna,
Desheng,
Elsa,
Kshitijaa,
Nicol\`{o},
Nori,
Robin,
Salima,
Stef,
and Zack), for creating the most incredible and wonderfully supportive environment.
I feel a deep sense of privilege to have been able to learn from you in these years.
    \item \hspace{-5mm} the DSSG team, for an absolutely incredible summer at CMU filled with inspiration, in particular, to Abby, Arun, and Catalina for the unique team spirit and friendship.
    
    \item \hspace{-5mm} the MPI-SF team (%
Ana,
André,
Celestine,
Florian,
Guanhua,
Jiduan,
Mila,
Mina,
Moritz,
Ricardo,
Vivian,
and Wale), for the brilliant exchange of ideas, the countless ping pong and volleyball games that made for an absolutely unforgettable time in Tübingen.
    
    \item \hspace{-5mm} the MilaNLP lab (%
Albert,
Arianna,
Debora,
Dirk,
Donya,
Elisa,
Flor,
Ema,
Paul,
Peppe,
and Tanise), for making the final part of my PhD remarkably enjoyable and for all the delightful ice cream breaks.
    
    \item \hspace{-5mm} my coauthors
Abby,
Aleksandra,
Alessandro,
Alice,
Andrea,
Daniele,
Anik\'{o},
Arun,
Catalina,
Celestine,
Christoph,
Corinna,
Desheng,
Eleonora,
Elsa,
Ezzat,
Giulia,
Kasun,
Kit,
Kshitijaa,
Markus,
Michele,
Nicole,
Nicol\`{o},
Piotr,
Rayid,
Riccardo,
Saverio,
Teresa,
Ulrich,
and Zack, for providing invaluable critical feedback and engaging in thought-provoking discussions that inspired creative solutions and a shared vision for how our machine learning research might truly change algorithmic systems for the better.
    
    \item \hspace{-5mm} Heidi, Mom, Dad, Sebi, Beni, Marta, and all my friends, for your love, unwavering support, and absolutely everything you have given me throughout this journey.
\end{itemize}

}

\clearpage

\tableofcontents
\clearpage

\pagenumbering{arabic}

\chapter{Introduction}

Imagine a future where AI-powered machines shape every aspect of human life.
Such machines determine who receives educational opportunities and who attends which schools, based on predicted academic success.
They evaluate job applications, selecting candidates most likely to succeed according to historical patterns of performance.
Financial decisions, such as loan approvals or insurance premiums, are calculated by algorithmic machines assessing risk profiles and economic potential.
Healthcare access is prioritized through AI models' predicted treatment outcomes and resource optimization.
Even social connections and online content consumption are curated by AI-powered recommendation systems that determine which relationships deserve attention and which content is worth consuming.
In this future world, the question arises of how we could evaluate if the AI's distribution of resources and opportunities is fair?
Would we comprehend the causal mechanisms and future directions our society might head toward as AI becomes increasingly integrated?
And how could we mitigate negative impacts on people's lives caused by unfavorable machine outcomes?
In an automated society where not only major decisions but also minor considerations in everyday life are taken by AI, questions about the societal impact of AI-powered machines would be fundamental.
Thus, AI's potential for unprecedented efficiency comes with the non-negligible risk of systematic disadvantages for less well-off socio-demographic groups.~\looseness=-1

This future is now.
While AI may not be fully responsible for all decision outcomes, our society is increasingly moving towards more efficient decision-making processes where more and more decisions are taken over by machine learning (ML).
Behind the human intermediaries or beneath familiar interfaces --- bank clerks, hiring managers, case workers, or job application tools --- ML predictions increasingly inform or make consequential decisions.
Already today, algorithmic systems scan resumes and predict job performance to shortlist candidates to be reviewed by human recruiters~\citep{Raghavan2020,Fabris2025FairnessSurvey}.
Loan officers present decisions that were largely predetermined by algorithms analyzing complex patterns beyond traditional credit scores~\citep{kozodoi2022credit-scoring,Fuster2022PredictablyMarkets}.
Insurance companies deploy personalized pricing models of which only the final outcomes are communicated to customers~\citep{Wuthrich2023,baumann2023PhilTech}.
Online platforms determine which housing or employment advertisements users see, potentially limiting opportunities~\citep{Ali2019}.
Criminal justice systems employ automated risk assessment tools to make potentially life-altering decisions about pretrial detention or sentencing~\citep{berk2021criminal}.

The core challenge of ML-powered decision systems lies in a fundamental misalignment between how ML systems are developed and what values society expects them to uphold.
ML models are typically optimized for prediction accuracy and operational efficiency rather than fairness or equity, creating systems that achieve their technical objectives without necessarily meeting broader societal expectations.
This misalignment, combined with the opacity surrounding many algorithmic decisions, creates an important area for research, as many people whose lives are significantly impacted by such systems remain unaware of ML's influence on their opportunities.
In theory, it would be possible to design ML-based systems specifically for social good, with goals and evaluation metrics that prioritize positive societal impact over other considerations~\citep{saleiro2018aequitas,hager2019artificial,tomavsev2020ai,rodolfa2020recidivsm,floridi2021design,baumann2024aaai}.
However, most commercial deployments pursue different objectives.
The fairness, accountability, and transparency (FAccT) research community\footnote{
See \url{https://facctconference.org/}
} has uncovered numerous cases of algorithmic discrimination across various domains, showcasing that questions about fairness and bias are not hypothetical future-looking intellectual pastimes --- they are urgent contemporary challenges.

This thesis confronts these issues through a thorough investigation of how ML-powered decision-making systems impact individuals and communities.
In particular, we address three interconnected goals that create a pathway toward fairness in ML for those affected by algorithmic decisions, especially members of historically disadvantaged or marginalized groups.
First, we introduce frameworks to measure fairness across different contexts, recognizing that appropriate metrics depend on domain-specific ethical considerations and the specific harms that different stakeholders might experience.
Second, we analyze the underlying mechanisms that create and amplify unfairness, examining both static bias and dynamic feedback effects.
Third, we develop practical solutions that mitigate ML-induced disparities while preserving their utility.
Together, these contributions advance our understanding of how to build more equitable algorithmic systems that distribute benefits and opportunities broadly, without marginalizing overlooked socio-demographic groups.

\section{Background}
\label{sec:Background}

\paragraph{Automated decision-making and recommender systems.}
Machine learning (ML) systems increasingly automate decisions that directly impact human lives and opportunities. These automated decision-making systems evaluate individuals based on historical data patterns, determining who receives loans~\citep{Fuster2022PredictablyMarkets}, insurance coverage~\citep{Wuthrich2023}, job interviews~\citep{Raghavan2020}, college admissions~\citep{Kleinberg2018}, or criminal risk assessments~\citep{berk2021criminal}.
Beyond such explicit decisions, recommender systems implicitly shape opportunity distribution and information access by determining which content users encounter online --- ranging from job advertisements~\citep{Ali2019} and housing opportunities~\citep{doj2022meta} to news articles and music recommendations~\citep{Jannach2023WhatRecommenders}.
This algorithmic mediation creates an environment where certain opportunities, information, or benefits may be systematically less accessible to specific demographic groups.

At the heart of these systems typically lies a pipeline that transforms data into decisions through ML-generated predictions based on individual characteristics.
Most decision-making scenarios, and all case studies and domains investigated in this thesis, can be mapped to a simplified structure as follows:
Suppose a decision maker (a firm or a platform) needs to make a binary decision $D \in \{0,1\}$ for each individual $i$.
This decision is based on a feature vector $\mathbf{x}_i$, containing known individual characteristics including a protected attribute $a_i \in A$ denoting group membership (sometimes also called \textit{sensitive} or \textit{protected attribute}).
In the classical supervised ML setting, the key variable guiding this process is a target variable $Y$, representing the ground truth outcome (e.g., loan repayment, recidivism, or job performance).
The fundamental challenge is that $Y$ is unknown at decision time.
In an idealized scenario, a perfect predictor would assign $\hat{Y}=Y$ for all individuals~\citep{murphy2012machine,mitchell2021algorithmic}.
However, since this is impossible in practice, ML systems use a prediction $\hat{Y} = f(X)$, which is learned from a set of known feature label pairs $(X,Y)$.
This estimation is usually imperfect for multiple reasons: the training data may contain measurement errors (where observed features $\mathbf{x}_i$ differ from the actual construct they represent), important variables may be omitted, and historical biases may be encoded in past decisions.
The prediction $\hat{Y}$ is used to inform a final decision $D$.
Thereby, the decision rule, which we denote by $r$, specifies how a decision is taken based on the individual prediction.
In its simplest form, $r$ applies a uniform threshold $\tau$ where $D_i = 1$ if $\hat{y}_i > \tau$ and $D_i = 0$ otherwise.

While these systems offer efficiency and scalability, their black-box nature often obscures how decisions are made and on what basis. Unlike human decision-makers who can be questioned or held accountable, algorithmic systems apply consistent patterns learned from historical data --- potentially perpetuating or amplifying existing social inequalities when deployed across large populations. These risks highlight the need for careful assessments of societal impact and appropriate interventions in ML-based decision-making pipelines.

\paragraph{Real-world cases of algorithmic discrimination.}
Concerns about ML's societal impact emerged following highly publicized cases of discriminatory outcomes in automated systems. These cases have demonstrated that ML models can systematically disadvantage marginalized groups, often reproducing or amplifying existing societal inequalities. 
To get a feeling for the magnitude and scope of this problem in today's world, consider the following cherry-picked examples from diverse domains.
\begin{itemize}
    \item One of the earliest and most prominent cases is the COMPAS recidivism prediction tool, used in the U.S. criminal justice system. A ProPublica investigation revealed that Black defendants were nearly twice as likely to be incorrectly classified as high-risk compared to white defendants~\citep{angwin2016machine, brennan_evaluating_2009}. Similarly, crime prediction systems such as PredPol were found to disproportionately target Black and Latino neighborhoods, reinforcing biased policing practices while largely avoiding white neighborhoods~\citep{sankin2021crime}.
    \item Bias in automated decision-making has also been documented in various other high-stakes domains, such as healthcare. A widely used algorithm designed to identify high-need patients favored white patients over Black patients because it used healthcare costs as a proxy for health needs --- a flawed design choice that resulted in fewer Black patients receiving necessary support~\citep{obermeyer2019dissecting}.
    \item Algorithmic bias has also caused severe harm in the public sector. In the Netherlands, an algorithmic fraud detection system for childcare benefits disproportionately targeted low-income families and minorities, leading to wrongful accusations and devastating financial consequences~\citep{geiger2021discriminatory}. A similar case occurred in Australia, where the ``robodebt'' system incorrectly calculated social security overpayments and pursued repayments from vulnerable citizens~\citep{Chowdhury2024}.
    \item Employment-related algorithms have raised similar concerns. Automated video interview systems have been shown to produce varying personality assessments based on superficial factors such as eyewear or background environment, raising doubts about their objectivity and susceptibility to visual biases~\citep{Harlan2021}. Austria's public employment service used an algorithm to score job seekers based on their employment potential, systematically assigning lower scores to women, older individuals, and disabled persons --- reducing their access to support programs~\citep{kayser2019austria}.
\end{itemize}
These real-world cases illustrate that algorithmic discrimination is not merely a theoretical concern but a pressing societal challenge with tangible consequences for affected individuals and communities. As algorithmic systems become more pervasive, measuring their societal impact and preventing adverse outcomes have become central goals for research and policy in ML.

For a more comprehensive review of discrimination risks in algorithmic systems across various domains, see \citet{orwat2019diskriminierungsrisiken}.

\paragraph{Group fairness criteria in ML.}
Concerns over discriminatory outcomes in algorithmic systems have led to the development of formal fairness definitions within the ML community.
In computer science, the concept of fairness typically refers to an equity assessment of decision making systems~\citep{barocas-hardt-narayanan}.
This thesis predominantly focuses on group fairness approaches, though other notions exist, for example individual fairness (measuring outcome differences across individuals that are similar according to a given distance metric~\citep{Dwork2012,Speicher2018individualandgroupfairness}) or causal definitions of fairness (e.g., counterfactual fairness~\citep{kusner2017counterfactual,Alvarez2023Counterfactual}).
The contextual importance of (group) fairness assessments impedes a straightforward application to the legal doctrine.
We refer the interested reader to philosophical~\citep{lippert-rasmussen_born_2014,Binns2018lessons,hertweck2021moral,baumann2022SDS_fairness_principle,Loi_Herlitz_Heidari_2024} and legal~\citep{10.2307/24758720,Hellman2020Measuring} literature for a broader discussion of terminological nuances.

Statistical fairness definitions test whether socially salient socio-demographic (sub)groups are treated equally.
Used as a criterion of ML's discriminatory impact, they can be used to minimize group-level disparities.
Three major categories are commonly discussed:
demographic parity (or statistical parity), which requires that the rate of positive decisions be equal across groups ($P(D=1 \mid A=a)=P(D=1 \mid A \neq a)$);
equalized odds (or separation), which ensures that error rates such as false positive and true positive rates are balanced ($P(D=1 \mid Y=i,A=a)=P(D=1 \mid Y=i,A \neq a), i \in \{0,1\}$);
and sufficiency, which requires predictions to have equal positive and negative predictive values across groups ($P(Y=1 \mid D=j,A=a)=P(Y=1 \mid D=j,A \neq a), j \in \{0,1\}$)~\citep{Dwork2012,hardt2016equality,Chouldechova2017,verma2018fairness}.
These criteria formalize different notions of fairness and often lead to incompatible requirements, as demonstrated by several impossibility theorems~\citep{Kleinberg2016,Chouldechova2017}.
Variants and extensions of these criteria, such as conditional statistical parity~\citep{Kamiran2013,10.1145/3097983.3098095} or predictive parity~\citep{Chouldechova2017}, offer additional nuance but still operate within a constrained design space of parity-based constraints.

Selecting the appropriate fairness criterion is not purely technical: it necessitates normative choices about which fairness notion to prioritize, reflecting normative judgments about what kind of inequality is acceptable in a given context~\citep{Heidari2019,Jacobs2021,baumann2022SDS_fairness_principle,Loi_Herlitz_Heidari_2024}.
Understanding these trade-offs is essential when evaluating or designing fair ML systems.
While these criteria are hard constraints that are either achieved or not, \textit{group fairness metrics} compare different groups, for example, by computing the difference or the ratio across groups' outcomes, and are helpful in that they quantify the extent to which such a criterion is fulfilled.

Despite their usefulness, group fairness criteria face several conceptual and operational limitations.
Most focus on parity in decision or error rates but remain agnostic to downstream consequences~\citep{Binns2018lessons}. They have also been criticized for the ``leveling down objection'', where enforcing equality can reduce outcomes for all groups~\citep{Parfit1995,mittelstadt2023unfairness}.
Trade-offs also arise between fairness and performance~\citep{hardt2016equality, 10.1145/3097983.3098095}, and between different fairness concepts such as group and individual fairness~\citep{Lipton2018Does,Binns2020}.

\paragraph{Dynamic aspects of fairness in ML systems.}
Most fairness criteria adopt a static perspective that fails to account for how ML systems impact their environment over time~\citep{Chouldechova2018,mitchell2021algorithmic,Liu2018Delayed_Long_version}.
The relationship between ML models and society is bidirectional and cyclical.
ML systems not only produce potentially unfair outcomes at a single point in time, but they can also fundamentally impact the social world their measurement is grounded in (represented by the red arrow). This impact in some cases extends to influencing the very outcomes they attempt to predict --- a phenomenon known as ``performative prediction''~\citep{Perdomo2020PerformativePrediction}.
For instance, predictive policing algorithms may direct increased surveillance toward certain neighborhoods, generating more arrests that further justify heightened policing, creating a self-fulfilling prophecy independent of actual crime rates \citep{Ensign2018RunawayPolicing}. However, the effect goes the other direction as well (blue arrow); society shapes ML systems through the data that humans generate, which often constitute the training data for the models, which learn from past patterns and outcomes.
This creates what has been called ``data leverage''~\citep{Vincent2021DataLeverage}.
Conscious data contribution provides users with a lever to promote their interests on digital platforms and counter existing power imbalances~\citep{Vincent2021ConsciousDataContribution}.
When individuals act in collectives and generate strategically data, they can significantly influence model behavior~\citep{hardt2023algorithmic}.~\looseness=-1

The reciprocal interaction between ML and society manifests as feedback loops (green arrows) that can reinforce, amplify, or occasionally mitigate bias over time.
Similar dynamics manifest in recommendation systems that shape user preferences \citep{Mansoury2020FeedbackLoop} and lending models that influence borrower behavior~\citep{Liu2018Delayed_Long_version}.
Addressing fairness in such dynamic contexts requires understanding not just the immediate impacts of algorithmic decisions but also their downstream consequences as systems reshape the environments they measure~\citep{Zhang2019GroupFairness,DAmour2020FairnessStudies,sunbackfire}.

\paragraph{The multidisciplinary nature of ML's societal impact.}
Studying ML's societal impact requires a multidisciplinary approach spanning technical, ethical, legal, and social dimensions.
Evaluating the normative implications of different fairness approaches, how algorithms interact with existing social structures and institutions, and how fairness definitions align with anti-discrimination laws and regulations goes beyond the technical tools computer scientists can provide~\citep{heidari2019moral,Binns2018,10.2307/24758720}.
This thesis primarily engages with the technical aspects of fairness in ML, focusing specifically on group fairness criteria, different types of bias, and mitigation strategies rather than on other dimensions within the broad topic of trustworthy AI such as privacy, transparency, or explainability.
By concentrating on these specific dimensions, we develop targeted solutions while acknowledging that comprehensive approaches to algorithmic fairness ultimately require integration across disciplinary boundaries and fairness paradigms.

\section{Three Goals Towards Fairness in ML}

Even highly accurate ML systems risk producing discriminatory outcomes.
Consider a seemingly perfect ML model that can predict recidivism with maximum accuracy based on historical data.
However, what if past judicial decisions have led to far fewer individuals from certain groups being released, affecting the representativeness of the training data?
What if police are less likely to catch or detain recidivists from different demographic groups, creating group-specific measurement errors in the outcomes to be predicted?
What if the ML prediction informs a decision for which future recidivism rates do not represent morally justifiable grounds for discriminatory treatment decisions?
In all these cases, improved predictive accuracy cannot solve the fundamental problem of unfairness.

Unfortunately, datasets or decision systems free of any bias hardly ever exist in practice.
ML-based decision systems are not fair by default.
They inherit and potentially amplify biases present in their training data and development processes.
In this thesis, we place people at the center of consideration and examine ML fairness as a distinct target, which has historically been neglected by the ML community in favor of optimization objectives like overall prediction accuracy.

ML models used in high-stakes decision making systems are often optimized for narrow technical objectives rather than broader social values, creating tensions between decision maker utility maximization and fairness considerations \citep{10.1145/3097983.3098095, hu2020fair}.
Current approaches to algorithmic fairness frequently address isolated aspects of the problem --- developing metrics without mechanisms for intervention, or implementing mitigations without understanding root causes.
This fragmentation limits the effectiveness of fairness efforts in real-world deployments.

Addressing these challenges requires engaging with three fundamental aspects of algorithmic fairness that naturally emerge when examining the full lifecycle of ML systems:
First, determining appropriate metrics to assess fairness that align with context-specific values;
Second, decomposing complex ML-based systems into meaningful components to systematically analyze mechanisms that generate and propagate bias.
And third, developing effective interventions that mitigate unfairness efficiently.
These three aspects are deeply interconnected.
Without appropriate measurement frameworks, we cannot identify fairness problems.
Without decomposing the system at hand, we cannot anticipate the impacts of fairness-enhancing interventions as the complexity of real-world decision making systems prohibits intuitive understanding of their dynamic behaviors and emergent properties.
And without practical interventions, even well-understood problems remain unresolved.
Each aspect represents a distinct research challenge while building upon insights from the others, forming a comprehensive approach to algorithmic fairness that spans technical implementation and normative considerations.

\subsection{{\color{goalIcolor}Measure}}

\paragraph{What's in a metric?}
The inherent quality of ML models that automate or assist decision-making systems is seldom capturable by a single metric.
Metrics transform abstract concepts into measurable quantities, but this transformation inevitably involves value judgments about what aspects matter most \citep{thomas2022reliance}.
Consider the study by~\citet{obermeyer2019dissecting} of a healthcare algorithm affecting millions of patients: by optimizing for healthcare costs rather than illness, the algorithm exhibited significant racial bias, identifying far fewer Black patients for additional care than if actual health needs had been measured.
Similarly, \citet{Mullainathan2017} found that apparent stroke risk factors in medical records were actually measuring healthcare utilization patterns rather than physiological risk.

Such cases hint at the ``metric fixation'' problem of ML systems~\citep{Muller2018Tyranny}.
Goodhart's Law applies here, as these systems optimize for given metrics, potentially at the expense of unmeasured values~\citep{Manheim2023Building}.

\paragraph{Measuring societal impact.}
Quantifying the societal impact of ML systems presents significant challenges.
Group fairness has emerged as one broadly accepted notion worth measuring because it is widely desirable from a societal perspective and due to its connection to non-discrimination laws.
Group fairness metrics translate abstract normative concepts into concrete mathematical formulations that can be measured and optimized in ML systems \citep{Corbett-Davies2018}.
Unlike traditional ML evaluation metrics like accuracy or precision that measure overall prediction quality, fairness metrics encode moral and ethical judgments about what constitutes equitable treatment across different groups in society~\citep{mitchell2021algorithmic,hertweck2021moral}.
However, evaluating only aggregate performance may obscure disparities across demographic groups.
Subgroup-specific analysis approaches have only recently become more popular, in particular in auditing work \citep{Buolamwini2018Gender,Raji2019Actionable}, and were later integrated into frameworks for standardized model reporting \citep{Mitchell2019Model}.
This shift from population-wide to subgroup-specific measurement represents an important evolution in how we measure ML systems' societal impacts, revealing disparities that population-level metrics hide and enabling more nuanced evaluation of fairness concerns.
\\\\
Going forward, we need well-defined, contextually appropriate metrics to evaluate whether ML systems produce equitable outcomes across different groups.
This recognition brings us to our first research goal:
\begin{tcolorbox}[colframe=goalIcolor!90, colback=goalIcolor!5, fonttitle=\bfseries, boxrule=1.5pt, left=3pt, right=3pt, top=1pt, bottom=3pt, boxsep=5pt]
\textit{\textbf{
\begin{enumerate}[left=0pt, topsep=2pt, itemsep=3pt, parsep=0pt, label={\color{goalIcolor}Goal \arabic* (Measure):}]
    \item {\color{goalIcolor}Establish appropriate metrics and frameworks to quantify fairness in ML systems across contexts.}
\end{enumerate}
}}
\end{tcolorbox}

\subsection{{\color{goalIIcolor}Decompose}}

While measuring fairness helps identify disparities, it tells us little about why these disparities occur or how they evolve over time.
Even appropriate fairness metrics cannot reveal the underlying structures through which bias enters and propagates through ML systems. This limitation points to our second goal:
\begin{tcolorbox}[colframe=goalIIcolor!90, colback=goalIIcolor!5, fonttitle=\bfseries, boxrule=1.5pt, left=3pt, right=3pt, top=1pt, bottom=3pt, boxsep=5pt]
\textit{\textbf{
\begin{enumerate}[left=0pt, topsep=2pt, itemsep=3pt, parsep=0pt, label={\color{goalIIcolor}Goal \arabic* (Decompose):}]
\setcounter{enumi}{1}
    \item {\color{goalIIcolor}Break down complex ML systems into their fundamental components to analyze bias dynamics and the emergence of unfairness over time.}
\end{enumerate}
}}
\end{tcolorbox}

\paragraph{Beyond symptom treatment.}
Breaking down complex ML-based decision systems is necessary for understanding the root causes of unfairness.
A meaningful decomposition is difficult yet essential for effective interventions.
When disparities emerge in algorithmic outcomes, organizations often implement quick fixes, which may temporarily reduce measured disparities while leaving intact the structures that generated the bias initially~\citep{Holstein2019improving}.
A deeper understanding of how bias enters and propagates through ML systems allows for structural approaches that address core problems rather than merely masking their effects \citep{Selbst2019fairness}.
This perspective shift from observing disparities to analyzing their origins is essential for developing meaningful solutions to algorithmic unfairness.

\paragraph{Bias can enter at multiple stages.}
Algorithmic bias emerges through various pathways throughout the ML development lifecycle~\citep{suresh2021framework}.
Bias may originate in historically discriminatory practices that shape the data we collect~\citep{Passi2019}, in sampling procedures that under-represent marginalized groups~\citep{Buolamwini2018Gender}, or in flawed variable operationalization that uses inadequate proxies for target concepts~\citep{obermeyer2019dissecting}.
Even with perfect data, bias can arise during model development through feature selection and algorithmic design choices that amplify subtle patterns~\citep{pedreschi2008discrimination-aware,Bolukbasi2016Man}.
The deployment context introduces yet another layer of potential bias when systems are used in environments different from those for which they were trained~\citep{Selbst2019fairness,suresh2021framework}.
Identifying precisely where and how bias enters the pipeline and what this means for downstream fairness enables equitable targeted interventions.

\paragraph{Static analysis misses dynamic effects.}
Approaches to algorithmic fairness mostly adopt a static perspective that fails to capture how systems evolve over time.
Many fairness audits examine model outputs at a single point in time, overlooking how initially small biases can magnify through feedback effects \citep{DAmour2020FairnessStudies}.
This limitation is particularly problematic for systems that make repeated decisions about individuals or populations, such as content recommendation systems, but also credit scoring models or predictive policing algorithms.
Decisions made today influence the environment that generates tomorrow's data, potentially creating self-reinforcing cycles of bias.
For example, a content recommendation system that slightly favors certain creators may generate a popularity gap that grows exponentially over time as popularity itself becomes a signal for quality \citep{Mansoury2020FeedbackLoop}.
Understanding these temporal dynamics requires moving beyond snapshot analyses to examine how fairness evolves as systems interact with their environments.

ML systems can create self-fulfilling prophecies through feedback effects that amplify initial biases.
When ML-based decisions influence future data collection or outcomes, they establish feedback loops that can dramatically increase disparities over time \citep{Ensign2018RunawayPolicing}.
Predictive policing algorithms direct increased surveillance toward certain neighborhoods, generating more arrests that further justify heightened policing --- regardless of actual crime rates.
Similarly, recommendation systems shape user preferences by determining what content people see, then treat those influenced preferences as ``ground truth'' for future recommendations \citep{chaney2018algorithmic}.
Credit scoring affects who receives loans, which influences repayment histories, which then affects future credit scores \citep{Liu2018Delayed_Long_version}.
These dynamic processes create complex challenges for fairness that static interventions cannot address.

\subsection{{\color{goalIIIcolor}Fix}}

Measuring fairness can reveal disparities, and effective decomposition of ML systems can shed light on the emergence and future trajectories of undesirable side effects like unfairness, but ultimately these insights must lead to action.
Without effective interventions, ML systems may continue to produce harmful impacts even when those impacts are well-documented and their mechanisms thoroughly analyzed.
This brings us to our third goal:
\begin{tcolorbox}[colframe=goalIIIcolor, colback=goalIIIcolor!5, fonttitle=\bfseries, boxrule=1.5pt, left=3pt, right=3pt, top=1pt, bottom=3pt, boxsep=5pt]
\textit{\textbf{
\begin{enumerate}[left=0pt, topsep=2pt, itemsep=3pt, parsep=0pt, label={\color{goalIIIcolor}Goal \arabic* (Fix):}]
\setcounter{enumi}{2}
    \item {\color{goalIIIcolor}Develop effective interventions that mitigate unfairness and contribute to a positive societal impact of ML.}
\end{enumerate}
}}
\end{tcolorbox}

\paragraph{Bias mitigation techniques.}
The ML literature offers a rich taxonomy of technical approaches to mitigate unfairness, typically categorized into three intervention points.
Pre-processing techniques transform the training data before model development, removing or reducing bias through methods like reweighting examples or transforming features \citep{kamiran2009classifying}.
In-processing approaches modify the learning algorithm itself, incorporating fairness constraints directly into the optimization objective \citep{Zafar2019, Agarwal2018}.
Post-processing is an intervention that changes the decision rule (see Chapter~\ref{sec:Background}) after model training~\citep{mehrabi2021survey}, which can adjust decisions without access to the internal structure of the prediction model, treating it as a black box.
Often, this results in applying different decision thresholds across groups \citep{hardt2016equality,baumann2022facct}.

Post-processing has proven to be a practical approach for mitigating unfairness~\citep{mitchell2021algorithmic}, but optimal rules under fairness constraints are only known for some metrics like demographic parity and equalized odds~\citep{hardt2016equality,pmlr-v81-menon18a}. These challenges highlight the need to combine technical tools with normative reasoning when operationalizing fairness in ML systems.
Each approach offers distinct trade-offs between effectiveness, implementation complexity, and compatibility with existing systems.
The choice among them depends on factors including access to the modeling pipeline, computational resources, and specific fairness requirements.

\paragraph{Collective action and data leverage.}
When direct modification of ML systems is not possible due to limited access or organizational resistance, external pressure through collective action offers an alternative pathway toward fairer outcomes.
Users, data subjects, and other stakeholders can leverage their collective data contributions to influence system behavior \citep{Vincent2021DataLeverage,Vincent2021ConsciousDataContribution}.
Strategic data contribution or withholding, such as coordinated efforts to modify training data patterns (referred to as \textit{algorithmic collective action} \citep{hardt2023algorithmic}), can shift model outputs in ways that promote underrepresented content or perspectives.
These approaches recognize that ML systems ultimately depend on data provided by users and communities, creating potential leverage points for those otherwise excluded from technical development processes.
Such collective strategies provide opportunities to drive positive societal impact when institutional incentives fail to prioritize them.

\paragraph{ML for social good.}
If developed responsibly, ML offers immense opportunities, even in high-stakes domains like social service distributions.
ML systems designed specifically for social good applications can allocate resources more effectively, identify individuals who could benefit from supportive interventions, or enhance public services \citep{saleiro2018aequitas,Rodolfa2021,hager2019artificial}.
Such applications require careful design to ensure they actually advance their stated goals without creating unintended consequences~\citep{tomavsev2020ai,rodolfa2020recidivsm,floridi2021design}.
Field validation becomes particularly important in these contexts, as theoretical benefits must translate to real-world improvements.

\section{Publications and Contributions}

This thesis presents eight papers that address the three goals outlined in the previous section. 

\textbf{\citet{baumann2023PhilTech} $\rightarrow$ Chapter~\ref{ssec:paper1}:}
\vspace{-1mm}
\begin{myquote}
\paperIcitation
\end{myquote}

\textbf{\citet{baumann2023EWAFgroup} $\rightarrow$ Chapter~\ref{ssec:paper2}:}
\vspace{-1mm}
\begin{myquote}
\paperIIcitation
\end{myquote}

\textbf{\citet{baumann2023facct} $\rightarrow$ Chapter~\ref{ssec:paper3}:}
\vspace{-1mm}
\begin{myquote}
\paperIIIcitation
\end{myquote}

\textbf{\citet{pagan2023eaamo} $\rightarrow$ Chapter~\ref{ssec:paper4}:}
\vspace{-1mm}
\begin{myquote}
\paperIVcitation
\end{myquote}

\textbf{\citet{baumann2024facct} $\rightarrow$ Chapter~\ref{ssec:paper5}:}
\vspace{-1mm}
\begin{myquote}
\paperVcitation
\end{myquote}

\textbf{\citet{baumann2022facct} $\rightarrow$ Chapter~\ref{ssec:paper6}:}
\vspace{-1mm}
\begin{myquote}
\paperVIcitation
\end{myquote}

\textbf{\citet{baumann2024NeurIPS} $\rightarrow$ Chapter~\ref{ssec:paper7}:}
\vspace{-1mm}
\begin{myquote}
\paperVIIcitation
\end{myquote}

\textbf{\citet{baumann2024aaai} $\rightarrow$ Chapter~\ref{ssec:paper8}:}
\vspace{-1mm}
\begin{myquote}
\paperVIIIcitation
\end{myquote}

\begin{figure}[p]
    \vspace{-5mm}
    \centering
    \includegraphics[width=1\textwidth]{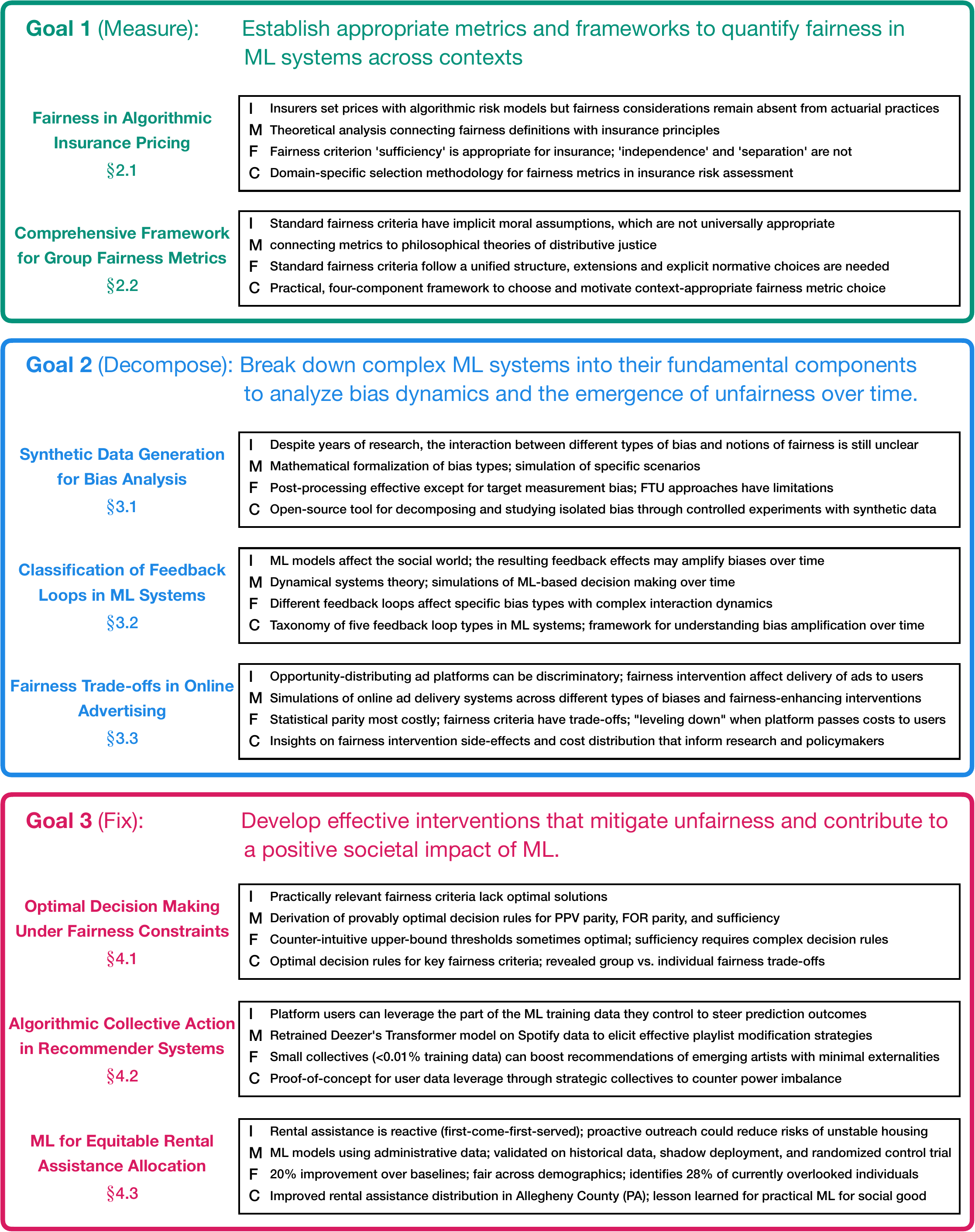}
    \caption{Summary of key ideas (\texttt{\textbf{I}}), methods (\texttt{\textbf{M}}), findings (\texttt{\textbf{F}}), and contributions (\texttt{\textbf{C}}) for the eight papers included in this thesis, organized by the three main research goals.}
    \label{fig:summary_of_contributions}
\end{figure}

\noindent
In these papers, various methods --- ranging from theoretical analysis and mathematical proofs to simulation studies and real-world implementation --- collectively advance our understanding of how to measure, decompose, and fix some of the potentially negative impacts ML systems might have on society.
Figure~\ref{fig:summary_of_contributions} presents an overview of all papers, highlighting each study's core idea, methodological approach, key findings, and contributions.
In the following sections, we briefly summarize each paper's key contributions and situate them within these three interconnected goals addressing the societal impact of ML.

\subsection*{{\color{goalIcolor}Goal 1: Measure}}

\subsubsection*{Fairness in Algorithmic Insurance Pricing \citep{baumann2023PhilTech} $\rightarrow \S$\ref{ssec:paper1}}
Insurance companies increasingly use algorithmic predictions to develop personalized risk models, raising important questions about fairness and discrimination~\citep{Dolman,Wuthrich2023}.
Prior work on algorithmic fairness has primarily focused on binary classification problems in domains like criminal justice or lending, leaving a gap in understanding fairness for non-binary outcomes like insurance pricing.
Our paper provides a thorough analysis of which fairness criteria are appropriate for insurance, examining the relationship between group fairness and actuarial fairness principles.
Through theoretical analysis connecting mathematical fairness definitions with normative insurance principles, we demonstrate that two commonly discussed criteria, independence (demographic parity) and separation (equalized odds), are not normatively appropriate for insurance premiums.
Instead, we find that another metric, sufficiency (also called well-calibration), is morally defensible in this context, allowing insurers to test for systematic biases in premiums across demographic groups.
We applied this framework to audit operational insurance products in Switzerland, contributing to a more equitable insurance pricing landscape.
This work enables insurance companies to assess the fairness properties of their risk models while providing a methodological blueprint for selecting appropriate fairness metrics based on domain-specific considerations.

\subsubsection*{A Comprehensive Framework for Group Fairness Metrics \citep{baumann2023EWAFgroup} $\rightarrow \S$\ref{ssec:paper2}}
Standard group fairness metrics have been increasingly criticized for their limitations: they focus exclusively on equality rather than other patterns of justice, they often target decision rates rather than actual impacts, and they provide limited guidance on selecting appropriate metrics for specific contexts~\citep{Feder2021emergent}.
Addressing these gaps, we develop a comprehensive framework that structures and extends standard fairness metrics by connecting them to theories of distributive justice.
Theories of distributive justice are concerned with the question of when the benefits and harms produced by decision systems can be called just~\citep{kuppler2021distributive,sep-justice-distributive}.
The framework consists of four components: defining utility for decision subjects, identifying relevant comparison groups, determining features that justify inequalities, and selecting appropriate patterns of justice.
Through theoretical analysis and case studies, we demonstrate how this framework reveals the normative choices implicit in technical implementations of fairness and allows for interpretation of their moral substance. Our approach enables practitioners to move beyond standard metrics toward context-appropriate fairness measures while making normative assumptions explicit.
This contribution supports more thoughtful selection of fairness metrics and helps assess the validity of fairness claims in audits and evaluations of algorithmic systems.

\subsection*{{\color{goalIIcolor}Goal 2: Decompose}}

\subsubsection*{Synthetic Data Generation for Bias Analysis \citep{baumann2023facct} $\rightarrow \S$\ref{ssec:paper3}}
Decomposing how different types of bias affect ML systems is challenging because real-world datasets often contain multiple intertwined biases.
To address this limitation, we develop a framework for generating synthetic datasets with controlled forms of bias, allowing for systematic isolation and analysis of individual biases.
By mathematically formalizing fundamental bias types (such as historical, measurement, representation, and omitted variable bias), we decompose the complex phenomenon of group-specific differences, enabling a systematic analysis of how these affect fairness metrics and model performance.
Experiments with our synthetic data generator confirm theoretical results and provide new insights:
post-processing techniques can effectively mitigate many bias types but struggle with measurement bias on target variables;
fairness criteria are often mutually exclusive as predicted by impossibility theorems;
and fairness through unawareness (FTU) approaches are limited.
Our open-source implementation allows researchers to decompose and study bias components that are difficult to isolate in real-world datasets, contributing tools and methodologies for more rigorous analysis of algorithmic fairness.

\subsubsection*{Classification of Feedback Loops in ML Systems \citep{pagan2023eaamo} $\rightarrow \S$\ref{ssec:paper4}}
ML systems deployed in real-world settings often create feedback loops that can amplify biases over time, yet prior work has typically relied on ad hoc models that prevent systematic comparison.
Our paper provides a formal definition and rigorous taxonomy of feedback loops in ML-based decision-making systems, breaking down complex dynamics using concepts from dynamical systems theory.
We decompose feedback loops into five distinct types based on which component of the decision-making pipeline is affected: sampling, individual, feature, outcome, and ML model feedback loops.
Through qualitative analysis and simulation examples in recommender systems, we examine how each feedback loop type affects biases throughout the ML pipeline.
This taxonomy covers many examples discussed in the algorithmic fairness literature, providing a unifying framework for studying dynamic bias propagation.
Our work shifts perspective from developing short-sighted solutions for existing biases toward a systematic breakdown of how biases emerge and amplify over time, creating foundations for more effective long-term fairness approaches.

\subsubsection*{Fairness Trade-offs in Online Advertising \citep{baumann2024facct} $\rightarrow \S$\ref{ssec:paper5}}
Online advertising platforms play a major role in distributing opportunities such as employment, housing, and credit, raising concerns about potential discrimination.
Despite a growing ability to measure unfair ad distribution, tools to model and predict how alternative systems might address these problems have been lacking. Our paper presents a simulation study modeling the effects of enforcing different fairness approaches in online advertising, revealing key trade-offs and unintended consequences.
We demonstrate that in realistic scenarios, 
pursuing fairness when one group is more expensive to advertise to often results in ``leveling down'' effects that benefit no group.
We also show that achieving statistical parity comes at a much higher utility cost to platforms than enforcing predictive parity or equality of opportunity.
Our analysis identifies a fundamental trade-off between different fairness notions: enforcing one criterion leads to worse outcomes on other criteria.
We find that negative effects can be prevented by ensuring platforms rather than users carry the fairness costs.
Our findings contribute to ongoing discussions on fair ad delivery by showing that fairness is not satisfied by default and that regulatory choices may backfire if potential side-effects are not considered.

\subsection*{{\color{goalIIIcolor}Goal 3: Fix}}

\subsubsection*{Optimal Decision Making Under Fairness Constraints \citep{baumann2022facct} $\rightarrow \S$\ref{ssec:paper6}}
Decision-making classifiers that maximize utility are not inherently fair, creating a need for decision rules that balance performance with fairness constraints.
While optimal decision rules have been derived for some fairness metrics, no such solutions existed for positive predictive value (PPV) parity, false omission rate (FOR) parity, and sufficiency. Our paper closes this gap by formulating algorithmic fairness as a constrained optimization problem and deriving optimal decision rules for these criteria. Through mathematical analysis and empirical evaluation, we show that, in most cases, group-specific threshold rules are optimal for PPV parity and FOR parity, similar to other fairness criteria.
However, surprisingly, we discover that in some cases, an upper-bound threshold rule for one group is optimal: selecting individuals with the smallest utility instead of the most promising ones.
This counterintuitive result contrasts with solutions for statistical parity and equality of opportunity.
For sufficiency, we demonstrate that more complex decision rules are required, leading to within-group unfairness for all but one group.
These findings highlight important trade-offs between group fairness and individual fairness that decision-makers must consider when implementing fairness constraints.

\subsubsection*{Collective Action in Recommender Systems \citep{baumann2024NeurIPS} $\rightarrow \S$\ref{ssec:paper7}}
Recommender systems on music streaming platforms often concentrate visibility on a small fraction of already-popular content, creating significant visibility barriers for emerging artists.
When platforms cannot be directly influenced to change their algorithms, alternative intervention strategies become necessary.
Our paper investigates algorithmic collective action as a mechanism for promoting underrepresented content, focusing on music streaming platforms where fans aim to increase visibility for emerging artists.
We introduce two easily implementable strategies for strategically placing songs in existing playlists to boost recommendations, exploiting statistical properties of transformer-based recommendation algorithms.
Evaluation using a publicly available model, productively deployed by the music streaming platform Deezer (where it affects millions of users), and user data from Spotify demonstrates that even small collectives (controlling less than 0.01\% of training data) can achieve up to 40$\times$ more recommendations than average songs with identical training frequency.
Our analysis of externalities shows that recommendations of other songs are largely preserved, suggesting that carefully designed collective action strategies can be effective without necessarily being adversarial.
This work demonstrates a viable intervention approach when direct system modification is not possible, providing empowering strategies for users affected by algorithmic systems.

\subsubsection*{ML for Equitable Rental Assistance Allocation \citep{baumann2024aaai} $\rightarrow \S$\ref{ssec:paper8}}
Rental assistance programs help prevent homelessness by providing financial support to individuals facing eviction, but these programs typically operate reactively and on a first-come, first-served basis rather than based on risk of future homelessness.
In collaboration with Allegheny County's Department of Human Services (ACDHS), Pennsylvania, we developed an ML-based system to proactively identify individuals at high risk of homelessness among those facing eviction.
Using county and state administrative data, we built predictive models that outperform simpler prioritization approaches by at least 20\% while remaining fair and equitable across race and gender.
Our proactive approach would identify 28\% of individuals who are overlooked by the current process and end up homeless.
Through extensive experiments with historical data and a shadow mode deployment, we validated our approach, demonstrating how ML can enhance resource allocation in social service contexts.
This work represents ML deployed specifically for social good, showing how fairness considerations can be integrated into assistive interventions while providing broader lessons for developing evidence-based decision support tools in similar contexts.

\paragraph{2025 update.}
Since the publication of our initial work \citep{baumann2024aaai}, the ML-based rental assistance system has moved from concept to active deployment in Allegheny County.
The system now generates weekly lists of approximately 30 individuals at high risk of homelessness for proactive outreach by ACDHS and their partner organizations.
We implemented a tailored outreach approach that routes individuals through appropriate channels based on their existing service connections (for example, families with active child welfare case managers receive outreach through those established relationships).
In early 2025, we launched a randomized controlled trial (RCT) to rigorously evaluate the effectiveness of our ML-based targeting as an augmentation to the current practice.
This evaluation design emerged after extensive discussions with ACDHS about feasible implementation options.
While the department remains committed to serving all individuals who proactively seek assistance, the addition of our ML-informed outreach creates an opportunity to reach those who might otherwise be overlooked.
The current implementation involves a weekly data pipeline where we fetch updates from the county data warehouse, generate predictions, and distribute targeted outreach lists to various partners.
We are now collaborating with ACDHS technical teams to transfer the models and operational processes to their infrastructure for long-term sustainability.
Funding uncertainty remains a challenge for the rental assistance program, and ACDHS plans to reassess the balance between reactive and proactive approaches as their resource situation evolves.
Despite these constraints, the deployment represents a significant step toward evidence-based, equity-centered resource allocation in public services.

\section{Conclusion}

This final section summarizes our key contributions, acknowledges limitations, and outlines directions for future research.
This thesis addressed three interconnected goals for understanding machine learning's (ML) societal impact: measuring fairness (Chapter~\ref{chaptergoal1}), decomposing complex mechanisms that generate bias (Chapter~\ref{chaptergoal2}), and developing effective interventions to drive a positive societal impact of ML systems (Chapter~\ref{chaptergoal3}).
Together, these contributions advance our understanding of how ML systems can distribute benefits and opportunities more equitably.

\paragraph{{\color{goalIcolor}Goal 1: Measure.}}

Determining appropriate metrics for fairness assessment requires domain-specific normative reasoning.
In the insurance domain, we developed a framework determining which fairness criteria are morally appropriate for algorithmic pricing
\citep{baumann2023PhilTech}. Through theoretical analysis connecting mathematical fairness definitions with insurance principles,
we identified sufficiency (well-calibration) as morally defensible in this context, while demonstrating that independence (demographic parity) and separation (equalized odds) are unsuitable.
We applied this framework to audit operational insurance products in Switzerland, contributing to a more equitable insurance pricing landscape.

Our comprehensive framework for group fairness metrics reveals the normative choices embedded in technical implementations.
By connecting fairness metrics to established philosophical theories of distributive justice
\citep{baumann2023EWAFgroup}, we unveiled the shared structure of group fairness criteria while making explicit the value judgments underlying different approaches.
The framework enables practitioners to interpret, extend, and unify group fairness metrics, facilitating more appropriate selections in specific contexts.

\paragraph{{\color{goalIIcolor}Goal 2: Decompose.}}

Our synthetic data generation framework enables isolating and analyzing specific bias types in ML systems.
By formally representing fundamental bias types, such as historical, measurement, representation, and omitted variable bias
\citep{baumann2023facct}, we facilitated systematic examination of how different biases affect fairness metrics and model performance.
This decomposition confirmed prior work on fairness-accuracy trade-offs and revealed important limitations of FTU approaches.

ML-based decision systems create feedback loops that can amplify biases over time.
Our formal taxonomy classifies five distinct types of feedback loops based on which component of the ML decision making pipeline is affected
\citep{pagan2023eaamo}.
This classification provides a unified framework for analyzing dynamic bias propagation, shifting perspective from short-term fixes toward understanding long-term consequences of algorithmic deployment.

Enforcing fairness in online advertising reveals complex trade-offs between stakeholder interests.
Our simulation study
\citep{baumann2024facct} demonstrated that statistical parity typically comes at a higher utility cost to platforms than predictive parity or equality of opportunity. We identified cases where pursuing fairness can result in ``leveling down'' effects that benefit no group, particularly when one group is more expensive to advertise to.

\paragraph{{\color{goalIIIcolor}Goal 3: Fix.}}

Optimal decision rules under fairness constraints can sometimes require counterintuitive approaches.
We derived mathematical solutions for maximizing utility while satisfying positive predictive value parity, false omission rate parity, and sufficiency
\citep{baumann2022facct}. In certain scenarios, an upper-bound threshold (selecting individuals with the lowest predicted scores) may be optimal.
These findings highlight important tensions between group fairness and individual fairness.

Algorithmic collective action enables users to counter power imbalances in recommendation systems.
Our research contributes to emerging literature on user data as a lever for users to promote their interests on digital platforms
\citep{baumann2024NeurIPS}. Even small collectives controlling minimal training data can significantly influence recommended items, effectively counteracting contemporary issues like popularity bias.
This creates new opportunities for participatory approaches when direct system modification is impossible.

In our collaboration with Allegheny County, we developed a system to proactively identify individuals at high risk of homelessness following eviction
\citep{baumann2024aaai}.
This showed that ML, when designed for equitable resource allocation, can help social workers allocate resources more effectively.
The system has moved to active deployment with an RCT underway to evaluate its effectiveness in augmenting current approaches.

While these works address various aspects of ML's societal impact, the common theme that knits them together is their focus on the consequential effects of ML systems on society, particularly for vulnerable populations.
Rather than relying on aggregate evaluations where minorities or disadvantaged individuals might fall through the cracks, these contributions explicitly account for various subgroups who might be disproportionately affected by algorithmic systems.

\paragraph{Limitations.}
This thesis primarily examines ML in decision-making and recommender systems, focusing specifically on group fairness approaches.
Other important aspects of ML's societal impact, or even other types of fairness, such as individual or counterfactual fairness, are beyond the scope of this thesis.
Despite its deep roots in non-discrimination principles, our group-specific approach is limited by its presupposition of having access to individuals' group membership information, which may not always be feasible in practice~\citep{Ashurst2023Fairness}.

The works presented in this thesis are tailored to specific problem scenarios and their generalizability may be limited.
This is due to the context-specific nature of ML's societal impact, which requires targeted interventions.
Post-processing fairness solutions ($\S$\ref{ssec:paper6}) presuppose the ability to adjust decision rules.
However, especially in cases where the ML model developer's goals deviate largely from societal desiderata, changing the way in which ML outcomes are transformed into decisions or recommendations may not be feasible.
In fact, this limitation motivated our work on algorithmic collective action, which overcomes this constraint by enabling users to influence systems without requiring direct access to decision mechanisms.
It is important to note that our proposed strategies to steer recommender systems ($\S$\ref{ssec:paper7}) are not universally optimal.
They target sequence-aware models like Transformers~\citep{NIPS2017_3f5ee243,Bendada2023DeezerPlaylistContinuationTransformers}, meaning alternative approaches would likely be more effective for sequence-unaware architectures.
Our rental assistance distribution system ($\S$\ref{ssec:paper8}) is optimized for Allegheny County's specific population and service context, though its methodological approach and the lessons learned could inform similar efforts elsewhere.

\paragraph{Future work.}
The fundamental fairness problems tackled in this thesis remain relevant across evolving ML architectures.
Our contributions are mostly model-agnostic, which means that the applicability of research findings goes beyond specific implementations.
Nevertheless, the landscape of ML usage has transformed significantly during the years in which the research presented in this thesis has been conducted.
As a result, continued research is needed to ensure a positive societal impact of ML.

Public access to powerful generative AI creates unprecedented equity challenges.
Unlike just a few years ago, the broader public now interacts directly with sophisticated AI systems through chatbots, image generators, and AI-enhanced search engines.
This widespread adoption introduces new opportunities and concerns.
For example, individuals with limited technical literacy or from underrepresented linguistic backgrounds may face barriers to accessing high-quality information through these systems~\citep{liu-etal-2023-evaluating,venkit2024search,Shah2024Envisioning}.
Scientific communities are simultaneously adopting ML models faster than we can understand their reliability~\citep{Park2023Generative,Argyle_Busby_Fulda_Gubler_Rytting_Wingate_2023,lu2024ai,yamada2025ai}, creating significant methodological challenges~\citep{Egami2023DSL,messeri2024artificial,wang2025large}.
Such rapid technological integration requires renewed scrutiny of societal implications across diverse contexts and user populations, as AI increasingly mediates access to critical resources and opportunities.

General-purpose AI systems demand new evaluation approaches~\citep{scantamburlo2024prediction}.
Unlike the dedicated decision and recommendation systems analyzed in this thesis, tools like large language models (LLMs) serve diverse, often user-determined applications that extend well beyond the use cases they were originally designed for~\citep{demszky2023using}.
However, as demonstrated throughout this thesis, ML systems do not produce equitable outputs by default.
For this reason, group fairness considerations remain relevant even in the era of general-purpose AI systems.
However, future evaluation techniques must account for how different socially salient groups interact with these systems and the consequential impacts of those interactions.
In general, evaluating increasingly complex ML systems remains challenging~\citep{sainz-etal-2023-nlp,deng-etal-2024-investigating,mirzadeh2025gsmsymbolic}.
As compound systems like retrieval-augmented LLMs become more prevalent~\citep{compound-ai-blog}, decomposing them into smaller components could help identify failure modes and appropriate safeguards.

Participatory approaches will be crucial for ensuring equitable benefits from ML's expanding role in society.
The generalizability of fairness frameworks to emerging technologies remains an open challenge, as does their integration into legal standards~\citep{Hellman2020Measuring,WACHTER2021105567,Weerts2023Algorithmic}.
Therefore, beyond multidisciplinary research, approaches that actively involve affected communities~\citep{Vincent2021DataLeverage,abeba22participatoryAI,hardt2023algorithmic,baumann2024NeurIPS} will grow increasingly important to ensure that ML's broadening adoption benefits society broadly rather than exacerbating existing inequalities.

\clearpage

\bibliographystyle{plainnat}
\bibliography{references}

@inproceedings{verma2018fairness,
author = {Verma, Sahil and Rubin, Julia},
booktitle = {Proceedings of the International Workshop on Software Fairness},
pages = {1--7},
series = {FairWare '18},
title = {{Fairness Definitions Explained}},
url = {https://doi.org/10.1145/3194770.3194776},
year = {2018}
}

@article{mehrabi2021survey,
address = {New York, NY, USA},
author = {Mehrabi, Ninareh and Morstatter, Fred and Saxena, Nripsuta and Lerman, Kristina and Galstyan, Aram},
journal = {ACM Comput. Surv.},
number = {6},
publisher = {Association for Computing Machinery},
title = {{A Survey on Bias and Fairness in Machine Learning}},
url = {https://doi.org/10.1145/3457607},
volume = {54},
year = {2021}
}

@inproceedings{baumann2022facct,
author = {Baumann, Joachim and Hann\'{a}k, Anik\'{o} and Heitz, Christoph},
title = {Enforcing Group Fairness in Algorithmic Decision Making: Utility Maximization Under Sufficiency},
year = {2022},
url = {https://doi.org/10.1145/3531146.3534645},
booktitle = {Proceedings of the 2022 ACM Conference on Fairness, Accountability, and Transparency},
pages = {2315–2326},
numpages = {12},
series = {FAccT '22}
}

@inproceedings{baumann2023facct,
author = {Baumann, Joachim and Castelnovo, Alessandro and Crupi, Riccardo and Inverardi, Nicole and Regoli, Daniele},
title = {Bias on Demand: A Modelling Framework That Generates Synthetic Data With Bias},
year = {2023},
url = {https://doi.org/10.1145/3593013.3594058},
booktitle = {Proceedings of the 2023 ACM Conference on Fairness, Accountability, and Transparency},
pages = {1002–1013},
numpages = {12},
series = {FAccT '23}
}

@article{baumann2023PhilTech,
  title={Fairness and risk: an ethical argument for a group fairness definition insurers can use},
  author={Baumann, Joachim and Loi, Michele},
  journal={Philosophy \& Technology},
  volume={36},
  number={3},
  pages={45},
  year={2023},
  url={https://doi.org/10.1007/s13347-023-00624-9}
}

@inproceedings{pagan2023eaamo,
author = {Pagan, Nicol\`{o} and Baumann, Joachim and Elokda, Ezzat and De Pasquale, Giulia and Bolognani, Saverio and Hann\'{a}k, Anik\'{o}},
title = {A Classification of Feedback Loops and Their Relation to Biases in Automated Decision-Making Systems},
year = {2023},
url = {https://doi.org/10.1145/3617694.3623227},
booktitle = {Proceedings of the 3rd ACM Conference on Equity and Access in Algorithms, Mechanisms, and Optimization},
articleno = {7},
numpages = {14},
series = {EAAMO '23}
}

@article{baumann2024aaai,
title={Preventing Eviction-Caused Homelessness through ML-Informed Distribution of Rental Assistance},
volume={38},
url={https://ojs.aaai.org/index.php/AAAI/article/view/30246},
number={20},
journal={Proceedings of the AAAI Conference on Artificial Intelligence},
author={Vajiac, Catalina and Frey, Arun and Baumann, Joachim and Smith, Abigail and Amarasinghe, Kasun and Lai, Alice and Rodolfa, Kit T. and Ghani, Rayid},
year={2024},
pages={22393-22400}
}

@inproceedings{baumann2024facct,
author = {Baumann, Joachim and Sapiezynski, Piotr and Heitz, Christoph and Hannak, Aniko},
title = {Fairness in Online Ad Delivery},
year = {2024},
address = {New York, NY, USA},
url = {https://doi.org/10.1145/3630106.3658980},
booktitle = {Proceedings of the 2024 ACM Conference on Fairness, Accountability, and Transparency},
pages = {1418–1432},
numpages = {15},
series = {FAccT '24}
}

@inproceedings{baumann2024NeurIPS,
 author = {Baumann, Joachim and Mendler-D\"{u}nner, Celestine},
 booktitle = {Advances in Neural Information Processing Systems},
 pages = {119123--119149},
 title = {Algorithmic Collective Action in Recommender Systems: Promoting Songs by Reordering Playlists},
 url = {https://proceedings.neurips.cc/paper_files/paper/2024/file/d79792543133425ff79513c147dc8881-Paper-Conference.pdf},
 volume = {37},
 year = {2024}
}

@article{baumann2023EWAFgroup,
author = {Baumann, Joachim and Hertweck, Corinna and Loi, Michele and Heitz, Christoph},
journal = {Proceedings of the 2nd European Workshop on Algorithmic Fairness (EWAF'23)},
title = {{Unification, Extension, and Interpretation of Group Fairness Metrics for ML-Based Decision-Making}},
year = {2023},
url={https://ceur-ws.org/Vol-3442/paper-23.pdf}
}

@incollection{sep-justice-distributive,
author = {Lamont, Julian and Favor, Christi},
booktitle = {The {Stanford} Encyclopedia of Philosophy},
edition = {{W}inter 2},
editor = {Zalta, Edward N},
howpublished = {\url{https://plato.stanford.edu/archives/win2017/entries/justice-distributive/}},
publisher = {Metaphysics Research Lab, Stanford University},
title = {{Distributive Justice}},
year = {2017}
}

@inproceedings{feder2021emergent,
address = {New York, NY, USA},
author = {Cooper, A Feder and Abrams, Ellen},
booktitle = {Proceedings of the 2021 AAAI/ACM Conference on AI, Ethics, and Society},
doi = {10.1145/3461702.3462519},
isbn = {9781450384735},
pages = {46--54},
publisher = {Association for Computing Machinery},
series = {AIES '21},
title = {{Emergent Unfairness in Algorithmic Fairness-Accuracy Trade-Off Research}},
url = {https://doi.org/10.1145/3461702.3462519},
year = {2021}
}

@inproceedings{rodolfa2020recidivsm,
address = {New York, NY, USA},
author = {Rodolfa, Kit T and Salomon, Erika and Haynes, Lauren and Mendieta, Iv{\'{a}}n Higuera and Larson, Jamie and Ghani, Rayid},
booktitle = {Proceedings of the 2020 Conference on Fairness, Accountability, and Transparency},
doi = {10.1145/3351095.3372863},
isbn = {9781450369367},
pages = {142--153},
publisher = {Association for Computing Machinery},
series = {FAT* '20},
title = {{Case Study: Predictive Fairness to Reduce Misdemeanor Recidivism through Social Service Interventions}},
url = {https://doi.org/10.1145/3351095.3372863},
year = {2020}
}

@article{obermeyer2019dissecting,
author = {Obermeyer, Ziad and Powers, Brian and Vogeli, Christine and Mullainathan, Sendhil},
journal = {Science},
number = {6464},
pages = {447--453},
publisher = {American Association for the Advancement of Science},
title = {{Dissecting racial bias in an algorithm used to manage the health of populations}},
volume = {366},
year = {2019}
}

@book{parfit1995,
author = {Parfit, Derek},
publisher = {Department of Philosophy, University of Kansas},
series = {The Lindley lecture},
title = {{Equality or priority}},
year = {1995}
}

@inproceedings{Binns2018lessons,
author = {Binns, Reuben},
booktitle = {Proceedings of the 1st Conference on Fairness, Accountability and Transparency},
editor = {Friedler, Sorelle A and Wilson, Christo},
pages = {149--159},
publisher = {PMLR},
series = {Proceedings of Machine Learning Research},
title = {{Fairness in Machine Learning: Lessons from Political Philosophy}},
url = {https://proceedings.mlr.press/v81/binns18a.html},
volume = {81},
year = {2018}
}

@article{kozodoi2022credit-scoring,
author = {Kozodoi, Nikita and Jacob, Johannes and Lessmann, Stefan},
doi = {https://doi.org/10.1016/j.ejor.2021.06.023},
issn = {0377-2217},
journal = {European Journal of Operational Research},
number = {3},
pages = {1083--1094},
title = {{Fairness in credit scoring: Assessment, implementation and profit implications}},
url = {https://www.sciencedirect.com/science/article/pii/S0377221721005385},
volume = {297},
year = {2022}
}

@book{lippert-rasmussen_born_2014,
address = {Oxford ; New York},
author = {Lippert-Rasmussen, Kasper},
isbn = {978-0-19-979611-3},
publisher = {Oxford University Press},
shorttitle = {Born free and equal?},
title = {{Born free and equal? a philosophical inquiry into the nature of discrimination}},
year = {2014}
}

@book{murphy2012machine,
author = {Murphy, Kevin P},
publisher = {MIT press},
title = {{Machine learning: a probabilistic perspective}},
year = {2012}
}

@article{Kamiran2013,
author = {Kamiran, Faisal and {\v{Z}}liobaitė, Indrė and Calders, Toon},
doi = {10.1007/s10115-012-0584-8},
issn = {0219-3116},
journal = {Knowledge and Information Systems},
number = {3},
pages = {613--644},
title = {{Quantifying explainable discrimination and removing illegal discrimination in automated decision making}},
url = {https://doi.org/10.1007/s10115-012-0584-8},
volume = {35},
year = {2013}
}

@inproceedings{pedreschi2008discrimination-aware,
address = {New York, NY, USA},
author = {Pedreschi, Dino and Ruggieri, Salvatore and Turini, Franco},
booktitle = {Proceedings of the 14th ACM SIGKDD International Conference on Knowledge Discovery and Data Mining},
doi = {10.1145/1401890.1401959},
isbn = {9781605581934},
pages = {560--568},
publisher = {Association for Computing Machinery},
series = {KDD '08},
title = {{Discrimination-Aware Data Mining}},
url = {https://doi.org/10.1145/1401890.1401959},
year = {2008}
}

@article{Ali2019,
address = {New York, NY, USA},
author = {Ali, Muhammad and Sapiezynski, Piotr and Bogen, Miranda and Korolova, Aleksandra and Mislove, Alan and Rieke, Aaron},
doi = {10.1145/3359301},
journal = {Proc. ACM Hum.-Comput. Interact.},
month = {nov},
number = {CSCW},
publisher = {Association for Computing Machinery},
title = {{Discrimination through Optimization: How Facebook's Ad Delivery Can Lead to Biased Outcomes}},
url = {https://doi.org/10.1145/3359301},
volume = {3},
year = {2019}
}

@article{Rodolfa2021,
author = {Rodolfa, Kit T and Lamba, Hemank and Ghani, Rayid},
doi = {10.1038/s42256-021-00396-x},
issn = {2522-5839},
journal = {Nature Machine Intelligence},
number = {10},
pages = {896--904},
title = {{Empirical observation of negligible fairness–accuracy trade-offs in machine learning for public policy}},
url = {https://doi.org/10.1038/s42256-021-00396-x},
volume = {3},
year = {2021}
}

@article{Agarwal2018,
author = {Agarwal, Alekh and Beygelzimer, Alina and Dud{\'{i}}k, Miroslav and Langford, John and Wallach, Hanna},
title = {{A Reductions Approach to Fair Classification}},
year = {2018}
}

@article{Zafar2019,
author = {Zafar, Muhammad Bilal and Valera, Isabel and Gomez-Rodriguez, Manuel and Gummadi, Krishna P},
journal = {Journal of Machine Learning Research},
number = {1},
pages = {2737----2778},
title = {{Fairness Constraints: A Flexible Approach for Fair Classification}},
url = {http://fate-computing.mpi-sws.org/.},
volume = {20},
year = {2019}
}

@misc{Corbett-Davies2018,
archivePrefix = {arXiv},
arxivId = {cs.CY/1808.00023},
author = {Corbett-Davies, Sam and Goel, Sharad},
eprint = {1808.00023},
primaryClass = {cs.CY},
title = {{The Measure and Mismeasure of Fairness: A Critical Review of Fair Machine Learning}},
url = {https://arxiv.org/abs/1808.00023},
year = {2018}
}

@inproceedings{pmlr-v81-menon18a,
address = {New York, NY, USA},
author = {Menon, Aditya Krishna and Williamson, Robert C},
booktitle = {Proceedings of the 1st Conference on Fairness, Accountability and Transparency},
editor = {Friedler, Sorelle A and Wilson, Christo},
pages = {107--118},
publisher = {PMLR},
series = {Proceedings of Machine Learning Research},
title = {{The cost of fairness in binary classification}},
url = {http://proceedings.mlr.press/v81/menon18a.html},
volume = {81},
year = {2018}
}

@inproceedings{Passi2019,
address = {New York, NY, USA},
author = {Passi, Samir and Barocas, Solon},
booktitle = {Proceedings of the Conference on Fairness, Accountability, and Transparency},
isbn = {9781450361255},
publisher = {ACM},
title = {{Problem Formulation and Fairness}},
url = {https://doi.org/10.1145/3287560.3287567},
year = {2019}
}

@article{Raghavan2020,
archivePrefix = {arXiv},
arxivId = {1906.09208},
author = {Raghavan, Manish and Barocas, Solon and Kleinberg, Jon and Levy, Karen},
doi = {10.1145/3351095.3372828},
eprint = {1906.09208},
journal = {FAT* 2020 - Proceedings of the 2020 Conference on Fairness, Accountability, and Transparency},
month = {jun},
pages = {469--481},
publisher = {Association for Computing Machinery, Inc},
title = {{Mitigating Bias in Algorithmic Hiring: Evaluating Claims and Practices}},
url = {http://arxiv.org/abs/1906.09208 http://dx.doi.org/10.1145/3351095.3372828},
year = {2020}
}

@article{Chouldechova2017,
author = {Chouldechova, Alexandra},
doi = {10.1089/big.2016.0047},
issn = {2167-647X (Electronic)},
journal = {Big data},
language = {eng},
month = {jun},
number = {2},
pages = {153--163},
pmid = {28632438},
title = {{Fair Prediction with Disparate Impact: A Study of Bias in Recidivism Prediction Instruments}},
volume = {5},
year = {2017}
}

@article{angwin2016machine,
author = {Angwin, Julia and Larson, Jeff and Mattu, Surya and Kirchner, Lauren},
journal = {ProPublica, May},
number = {2016},
pages = {139--159},
title = {{Machine bias}},
url = {https://www.propublica.org/article/machine-bias-risk-assessments-in-criminal-sentencing},
volume = {23},
year = {2016}
}

@inproceedings{lipton2018does,
archivePrefix = {arXiv},
arxivId = {1711.07076},
author = {Lipton, Zachary C. and Chouldechova, Alexandra and McAuley, Julian},
booktitle = {Advances in Neural Information Processing Systems},
eprint = {1711.07076},
issn = {10495258},
pages = {8136--8146},
publisher = {Curran Associates, Inc.},
title = {{Does mitigating ML's impact disparity require treatment disparity?}},
url = {https://proceedings.neurips.cc/paper/2018/file/8e0384779e58ce2af40eb365b318cc32-Paper.pdf},
volume = {31},
year = {2018}
}

@inproceedings{Jacobs2021,
address = {New York, NY, USA},
archivePrefix = {arXiv},
arxivId = {1912.05511},
author = {Jacobs, Abigail Z. and Wallach, Hanna},
booktitle = {FAccT 2021 - Proceedings of the 2021 ACM Conference on Fairness, Accountability, and Transparency},
doi = {10.1145/3442188.3445901},
eprint = {1912.05511},
isbn = {9781450383097},
month = {mar},
number = {21},
pages = {375--385},
publisher = {Association for Computing Machinery, Inc},
title = {{Measurement and fairness}},
url = {https://dl.acm.org/doi/10.1145/3442188.3445901},
volume = {11},
year = {2021}
}

@techreport{Binns2018,
author = {Binns, Reuben},
booktitle = {Proceedings of Machine Learning Research},
issn = {2640-3498},
month = {jan},
pages = {1--11},
publisher = {PMLR},
title = {{Fairness in Machine Learning: Lessons from Political Philosophy}},
url = {http://proceedings.mlr.press/v81/binns18a.html},
volume = {81},
year = {2018}
}

@inproceedings{Binns2020,
address = {New York, NY, USA},
archivePrefix = {arXiv},
arxivId = {1912.06883},
author = {Binns, Reuben},
booktitle = {FAT* 2020 - Proceedings of the 2020 Conference on Fairness, Accountability, and Transparency},
doi = {10.1145/3351095.3372864},
eprint = {1912.06883},
isbn = {9781450369367},
month = {jan},
pages = {514--524},
publisher = {Association for Computing Machinery, Inc},
title = {{On the apparent conflict between individual and group fairness}},
url = {https://dl.acm.org/doi/10.1145/3351095.3372864},
year = {2020}
}

@inproceedings{Dolman,
title = {{Algorithmic Fairness: Contemporary Ideas in the Insurance Context}},
url={https://www.actuaries.org.uk/system/files/field/document/B9_Chris%20Dolman%20%28paper%29.pdf},
  author={Dolman, Chris and Semenovich, Dimitri},
  booktitle={GIRO Conference},
  year={2018}
}

@article{Chouldechova2018,
archivePrefix = {arXiv},
arxivId = {1810.08810},
author = {Chouldechova, Alexandra and Roth, Aaron},
eprint = {1810.08810},
journal = {arXiv},
month = {oct},
publisher = {arXiv},
title = {{The Frontiers of Fairness in Machine Learning}},
url = {http://arxiv.org/abs/1810.08810},
year = {2018}
}

@unpublished{Kleinberg2016,
archivePrefix = {arXiv},
arxivId = {1609.05807v2},
author = {Kleinberg, Jon and Mullainathan, Sendhil and Raghavan, Manish},
eprint = {1609.05807v2},
title = {{Inherent Trade-Offs in the Fair Determination of Risk Scores}},
year = {2016}
}

@inproceedings{Dwork2012,
author = {Dwork, Cynthia and Hardt, Moritz and Pitassi, Toniann and Reingold, Omer and Zemel, Richard},
booktitle = {ITCS 2012 - Innovations in Theoretical Computer Science Conference},
pages = {214--226},
title = {{Fairness through awareness}},
url = {http://dl.acm.org/citation.cfm?doid=2090236.2090255},
year = {2012}
}

@article{berk2021criminal,
author = {Berk, Richard and Heidari, Hoda and Jabbari, Shahin and Kearns, Michael and Roth, Aaron},
journal = {Sociological Methods \& Research},
number = {1},
pages = {3--44},
title = {{Fairness in Criminal Justice Risk Assessments: The State of the Art}},
url = {https://doi.org/10.1177/0049124118782533},
volume = {50},
year = {2021}
}

@article{mitchell2021algorithmic,
author = {Mitchell, Shira and Potash, Eric and Barocas, Solon and D'Amour, Alexander and Lum, Kristian},
journal = {Annual Review of Statistics and Its Application},
number = {1},
pages = {141--163},
publisher = {Annual Reviews},
title = {{Algorithmic Fairness: Choices, Assumptions, and Definitions}},
url = {https://www.annualreviews.org/doi/10.1146/annurev-statistics-042720-125902},
volume = {8},
year = {2021}
}

@inproceedings{Bendada2023DeezerPlaylistContinuationTransformers,
author = {Bendada, Walid and Salha-Galvan, Guillaume and Bouab{\c{c}}a, Thomas and Cazenave, Tristan},
booktitle = {International ACM SIGIR Conference on Research and Development in Information Retrieval},
pages = {464--474},
title = {{A Scalable Framework for Automatic Playlist Continuation on Music Streaming Services}},
year = {2023}
}

@inproceedings{hardt2023algorithmic,
author = {Hardt, Moritz and Mazumdar, Eric and Mendler-D{\"{u}}nner, Celestine and Zrnic, Tijana},
booktitle = {International Conference on Machine Learning},
pages = {12570--12586},
title = {{Algorithmic Collective Action in Machine Learning}},
volume = {202},
year = {2023}
}

@article{Vincent2021ConsciousDataContribution,
author = {Vincent, Nicholas and Hecht, Brent},
journal = {Proc. ACM Hum.-Comput. Interact.},
title = {{Can "Conscious Data Contribution" Help Users to Exert "Data Leverage" Against Technology Companies?}},
volume = {5},
year = {2021}
}

@inproceedings{Vincent2021DataLeverage,
author = {Vincent, Nicholas and Li, Hanlin and Tilly, Nicole and Chancellor, Stevie and Hecht, Brent},
booktitle = {ACM Conference on Fairness, Accountability, and Transparency},
pages = {215--227},
title = {{Data Leverage: A Framework for Empowering the Public in Its Relationship with Technology Companies}},
year = {2021}
}

@InProceedings{Jannach2023WhatRecommenders,
author="Jannach, Dietmar
and Lerche, Lukas
and Gedikli, Fatih
and Bonnin, Geoffray",
title="What Recommenders Recommend -- An Analysis of Accuracy, Popularity, and Sales Diversity Effects",
booktitle="User Modeling, Adaptation, and Personalization",
year="2013",
publisher="Springer Berlin Heidelberg",
pages="25--37"
}

@inproceedings{abeba22participatoryAI,
author = {Birhane, Abeba and Isaac, William and Prabhakaran, Vinodkumar and Diaz, Mark and Elish, Madeleine Clare and Gabriel, Iason and Mohamed, Shakir},
title = {Power to the People? Opportunities and Challenges for Participatory AI},
year = {2022},
booktitle = {ACM Conference on Equity and Access in Algorithms, Mechanisms, and Optimization},
articleno = {6},
numpages = {8}
}

@article{saleiro2018aequitas,
  title={Aequitas: A bias and fairness audit toolkit},
  author={Saleiro, Pedro and Kuester, Benedict and Hinkson, Loren and London, Jesse and Stevens, Abby and Anisfeld, Ari and Rodolfa, Kit T and Ghani, Rayid},
  journal={arXiv preprint arXiv:1811.05577},
  year={2018}
}

@inproceedings{hardt2016equality,
 author = {Hardt, Moritz and Price, Eric and Price, Eric and Srebro, Nati},
 booktitle = {Advances in Neural Information Processing Systems},
 publisher = {Curran Associates, Inc.},
 title = {Equality of Opportunity in Supervised Learning},
 volume = {29},
 year = {2016}
}

@article{kusner2017counterfactual,
  title={Counterfactual fairness},
  author={Kusner, Matt J and Loftus, Joshua R and Russell, Chris and Silva, Ricardo},
  journal={arXiv preprint arXiv:1703.06856},
  year={2017}
}

@inproceedings{hertweck2021moral,
author = {Hertweck, Corinna and Heitz, Christoph and Loi, Michele},
title = {On the Moral Justification of Statistical Parity},
year = {2021},
url = {https://doi.org/10.1145/3442188.3445936},
booktitle = {Proceedings of the 2021 ACM Conference on Fairness, Accountability, and Transparency},
pages = {747–757},
numpages = {11},
series = {FAccT '21}
}

@inproceedings{heidari2019moral,
  title={A moral framework for understanding fair ML through economic models of equality of opportunity},
  author={Heidari, Hoda and Loi, Michele and Gummadi, Krishna P and Krause, Andreas},
  booktitle={Proceedings of the Conference on Fairness, Accountability, and Transparency},
  pages={181--190},
  year={2019}
}

@inproceedings{buolamwini2018gender,
  title={Gender shades: Intersectional accuracy disparities in commercial gender classification},
  author={Buolamwini, Joy and Gebru, Timnit},
  booktitle={Conference on fairness, accountability and transparency},
  pages={77--91},
  year={2018},
  organization={PMLR}
}

@misc{kuppler2021distributive,
      title={Distributive Justice and Fairness Metrics in Automated Decision-making: How Much Overlap Is There?}, 
      author={Matthias Kuppler and Christoph Kern and Ruben L. Bach and Frauke Kreuter},
      year={2021},
      eprint={2105.01441},
      archivePrefix={arXiv},
      primaryClass={stat.ML}
}

@inproceedings{selbst2019fairness,
  title={Fairness and abstraction in sociotechnical systems},
  author={Selbst, Andrew D and Boyd, Danah and Friedler, Sorelle A and Venkatasubramanian, Suresh and Vertesi, Janet},
  booktitle={Proceedings of the conference on fairness, accountability, and transparency},
  pages={59--68},
  year={2019}
}

@inproceedings{hu2020fair,
  title={Fair classification and social welfare},
  author={Hu, Lily and Chen, Yiling},
  booktitle={Proceedings of the 2020 Conference on Fairness, Accountability, and Transparency},
  pages={535--545},
  year={2020}
}

@inproceedings{baumann2022SDS_fairness_principle,
author = {Baumann, Joachim and Heitz, Christoph},
booktitle = {2022 9th Swiss Conference on Data Science (SDS)},
url = {https://doi.org/10.1109/SDS54800.2022.00011},
pages = {19--25},
title = {{Group Fairness in Prediction-Based Decision Making: From Moral Assessment to Implementation}},
year = {2022}
}

@article{10.2307/24758720,
author = {Barocas, Solon and Selbst, Andrew D},
journal = {California Law Review},
number = {3},
pages = {671--732},
publisher = {California Law Review, Inc.},
title = {{Big Data's Disparate Impact}},
url = {http://www.jstor.org/stable/24758720},
volume = {104},
year = {2016}
}

@inproceedings{Speicher2018individualandgroupfairness,
author = {Speicher, Till and Heidari, Hoda and Grgic-Hlaca, Nina and Gummadi, Krishna P. and Singla, Adish and Weller, Adrian and Zafar, Muhammad Bilal},
title = {A Unified Approach to Quantifying Algorithmic Unfairness: Measuring Individual \& Group Unfairness via Inequality Indices},
year = {2018},
url = {https://doi.org/10.1145/3219819.3220046},
booktitle = {Proceedings of the 24th ACM SIGKDD International Conference on Knowledge Discovery \& Data Mining},
pages = {2239–2248},
numpages = {10},
series = {KDD '18}
}

@inproceedings{Kamiran2009Classifying,
author = {Kamiran, Faisal and Calders, Toon},
booktitle = {2009 2nd International Conference on Computer, Control and Communication},
doi = {10.1109/IC4.2009.4909197},
pages = {1--6},
title = {{Classifying without discriminating}},
year = {2009}
}

@article{baumann2022distributive,
  title={Distributive justice as the foundational premise of fair ML: Unification, extension, and interpretation of group fairness metrics},
  author={Baumann, Joachim and Hertweck, Corinna and Loi, Michele and Heitz, Christoph},
  journal={arXiv preprint arXiv:2206.02897},
  year={2022},
note={Presented as a poster at the \textit{3rd ACM Conference on Equity and Access in Algorithms, Mechanisms, and Optimization (EAAMO'2023)}}
}

@article{brennan_evaluating_2009,
	title = {Evaluating the predictive validity of the compas risk and needs assessment system},
	volume = {36},
	number = {1},
	journal = {Criminal Justice and Behavior},
	author = {Brennan, Tim and Dieterich, William and Ehret, Beate},
	year = {2009},
	pages = {21--40},
}

@article{scantamburlo2024prediction,
  title={On prediction-modelers and decision-makers: why fairness requires more than a fair prediction model},
  author={Scantamburlo, Teresa and Baumann, Joachim and Heitz, Christoph},
  journal={AI \& SOCIETY},
  pages={1--17},
  year={2024},
  publisher={Springer},
  url={https://doi.org/10.1007/s00146-024-01886-3}
}

@book{barocas-hardt-narayanan,
  title = {Fairness and Machine Learning: Limitations and Opportunities},
  author = {Solon Barocas and Moritz Hardt and Arvind Narayanan},
  publisher = {MIT Press},
  year = {2023}
}

@inproceedings{suresh2021framework,
author = {Suresh, Harini and Guttag, John},
title = {A Framework for Understanding Sources of Harm throughout the Machine Learning Life Cycle},
year = {2021},
isbn = {9781450385534},
publisher = {Association for Computing Machinery},
address = {New York, NY, USA},
url = {https://doi.org/10.1145/3465416.3483305},
doi = {10.1145/3465416.3483305},
booktitle = {Equity and Access in Algorithms, Mechanisms, and Optimization},
articleno = {17},
numpages = {9},
location = {--, NY, USA},
series = {EAAMO '21}
}

@book{orwat2019diskriminierungsrisiken,
  title={Diskriminierungsrisiken durch Verwendung von Algorithmen: eine Studie, erstellt mit einer Zuwendung der Antidiskriminierungsstelle des Bundes},
  author={Orwat, Carsten},
  year={2019},
  publisher={Nomos}
}

@article{sankin2021crime,
  title={Crime Prediction Software Promised to Be Free of Biases. New Data Shows It Perpetuates Them},
  author={Sankin, Aaron and Mehrota, Dhruv and Mattu, Surya and Gilbertson, Annie},
  journal={The Markup},
  volume={2},
  year={2021}
}

@article{Harlan2021,
  title={OBJECTIVE OR BIASED: On the questionable use of Artificial Intelligence for job applications},
  author={Harlan, Elisa and Schnuck, Oliver},
  journal={{Bayerischer Rundfunk}},
  year={2021}
}

@article{geiger2021discriminatory,
  title={How a Discriminatory Algorithm Wrongly Accused Thousands of Families of Fraud},
  author={Geiger, Gabriel},
  journal={Vice},
  year={2021}
}

@article{kayser2019austria,
  title={Austria’s employment agency rolls out discriminatory algorithm, sees no problem},
  author={Kayser-Bril, Nicolas},
  journal={Algorithm Watch},
  volume={6},
  year={2019}
}

@article{Chowdhury2024,
author = {Saabiq Chowdhury},
title = {Technology is never neutral: Robodebt and a human rights analysis of automated decision-making on welfare recipients},
journal = {Australian Journal of Human Rights},
volume = {0},
number = {0},
pages = {1--21},
year = {2024},
publisher = {Routledge},
doi = {10.1080/1323238X.2024.2409620},
URL = {https://doi.org/10.1080/1323238X.2024.2409620},
eprint = {https://doi.org/10.1080/1323238X.2024.2409620}
}

@article{Heidari2019,
    title = {{A Moral Framework for Understanding Fair ML through Economic Models of Equality of Opportunity}},
    year = {2019},
    author = {Heidari, Hoda and Loi, Michele and Gummadi, Krishna P and Krause, Andreas},
    pages = {181--190},
    isbn = {9781450361255}
}

@inproceedings{10.1145/3097983.3098095,
    title = {{Algorithmic Decision Making and the Cost of Fairness}},
    year = {2017},
    booktitle = {Proceedings of the 23rd ACM SIGKDD International Conference on Knowledge Discovery and Data Mining},
    author = {Corbett-Davies, Sam and Pierson, Emma and Feller, Avi and Goel, Sharad and Huq, Aziz},
    pages = {797–806},
    series = {KDD '17},
    url = {https://doi.org/10.1145/3097983.3098095},
}

@article{Kleinberg2018,
    title = {{Algorithmic Fairness}},
    year = {2018},
    journal = {AEA Papers and Proceedings},
    author = {Kleinberg, Jon and Ludwig, Jens and Mullainathan, Sendhil and Rambachan, Ashesh},
    month = {5},
    pages = {22--27},
    volume = {108},
    url = {https://www.aeaweb.org/articles?id=10.1257/pandp.20181018},
}

@inproceedings{Liu2018Delayed_Long_version,
    title = {{Delayed Impact of Fair Machine Learning}},
    year = {2018},
    booktitle = {Proceedings of the 35th International Conference on Machine Learning},
    author = {Liu, Lydia T and Dean, Sarah and Rolf, Esther and Simchowitz, Max and Hardt, Moritz},
    editor = {Dy, Jennifer and Krause, Andreas},
    pages = {3150--3158},
    series = {Proceedings of Machine Learning Research},
    volume = {80},
    publisher = {PMLR},
    url = {https://proceedings.mlr.press/v80/liu18c.html}
}

@article{DAmour2020FairnessStudies,
    title = {{Fairness is not static: Deeper understanding of long term fairness via simulation studies}},
    year = {2020},
    journal = {FAT* 2020 - Proceedings of the 2020 Conference on Fairness, Accountability, and Transparency},
    author = {D'Amour, Alexander and Srinivasan, Hansa and Atwood, James and Baljekar, Pallavi and Sculley, D. and Halpern, Yoni},
    pages = {525--534},
    isbn = {9781450369367},
    doi = {10.1145/3351095.3372878}
}

@article{Zhang2019GroupFairness,
    title = {{Group retention when using machine learning in sequential decision making: The interplay between user dynamics and fairness}},
    year = {2019},
    journal = {Advances in Neural Information Processing Systems},
    author = {Zhang, Xueru and Khalili, Mohammad Mahdi and Tekin, Cem and Liu, Mingyan},
    number = {NeurIPS},
    volume = {32},
    issn = {10495258},
    arxivId = {1905.00569}
}

@article{chaney2018algorithmic,
    title = {{How algorithmic confounding in recommendation systems increases homogeneity and decreases utility}},
    year = {2018},
    author = {Chaney, Allison J. B. and Stewart, Brandon M. and Engelhardt, Barbara E.},
    pages = {224--232},
    isbn = {9781450359016},
    doi = {10.1145/3240323.3240370},
    arxivId = {1710.11214}
}

@article{Perdomo2020PerformativePrediction,
    title = {{Performative prediction}},
    year = {2020},
    journal = {37th International Conference on Machine Learning, ICML 2020},
    author = {Perdomo, Juan C. and Zrnic, Tijana and Mendler-Dunner, Celestine and Hardt, Moritz},
    pages = {7555--7565},
    volume = {PartF16814},
    isbn = {9781713821120},
    arxivId = {2002.06673}
}

@article{Fuster2022PredictablyMarkets,
    title = {{Predictably Unequal? The Effects of Machine Learning on Credit Markets}},
    year = {2022},
    journal = {Journal of Finance},
    author = {Fuster, Andreas and Goldsmith-Pinkham, Paul and Ramadorai, Tarun and Walther, Ansgar},
    number = {1},
    pages = {5--47},
    volume = {77},
    doi = {10.1111/jofi.13090},
    issn = {15406261}
}

@inproceedings{Ensign2018RunawayPolicing,
    title = {{Runaway Feedback Loops in Predictive Policing}},
    year = {2018},
    booktitle = {Proceedings of Machine Learning Research},
    author = {Ensign, Danielle and Friedler, Sorelle A and Neville, Scott and Scheidegger, Carlos and Venkatasubramanian, Suresh and Wilson, Christo},
    pages = {1--12},
    volume = {81},
    url = {https://github.com/algofairness/}
}

@article{sunbackfire,
    title = {{The Backfire Effects of Fairness Constraints}},
    year = {2022},
    journal = {ICML 2022 Workshop on Responsible Decision Making in Dynamic Environments},
    author = {Sun, Yi and Cuesta-Infante, Alfredo and Veeramachaneni, Kalyan},
    url = {https://responsibledecisionmaking.github.io/assets/pdf/papers/44.pdf}
}

@article{mittelstadt2023unfairness,
  title={The Unfairness of Fair Machine Learning: Leveling Down and Strict Egalitarianism by Default},
  author={Mittelstadt, Brent and Wachter, Sandra and Russell, Chris},
  journal={Michigan Technology Law Review},
  volume={30},
  pages={1},
  year={2023},
  publisher={HeinOnline}
}

@article{thomas2022reliance,
  title={Reliance on metrics is a fundamental challenge for AI},
  author={Thomas, Rachel L and Uminsky, David},
  journal={Patterns},
  volume={3},
  number={5},
  year={2022},
  publisher={Elsevier}
}

@inproceedings{Mitchell2019Model,
author = {Mitchell, Margaret and Wu, Simone and Zaldivar, Andrew and Barnes, Parker and Vasserman, Lucy and Hutchinson, Ben and Spitzer, Elena and Raji, Inioluwa Deborah and Gebru, Timnit},
title = {Model Cards for Model Reporting},
year = {2019},
isbn = {9781450361255},
publisher = {Association for Computing Machinery},
address = {New York, NY, USA},
url = {https://doi.org/10.1145/3287560.3287596},
doi = {10.1145/3287560.3287596},
booktitle = {Proceedings of the Conference on Fairness, Accountability, and Transparency},
pages = {220–229},
numpages = {10},
location = {Atlanta, GA, USA},
series = {FAT* '19}
}

@inproceedings{Bolukbasi2016Man,
 author = {Bolukbasi, Tolga and Chang, Kai-Wei and Zou, James Y and Saligrama, Venkatesh and Kalai, Adam T},
 booktitle = {Advances in Neural Information Processing Systems},
 editor = {D. Lee and M. Sugiyama and U. Luxburg and I. Guyon and R. Garnett},
 pages = {},
 publisher = {Curran Associates, Inc.},
 title = {Man is to Computer Programmer as Woman is to Homemaker? Debiasing Word Embeddings},
 url = {https://proceedings.neurips.cc/paper_files/paper/2016/file/a486cd07e4ac3d270571622f4f316ec5-Paper.pdf},
 volume = {29},
 year = {2016}
}

@book{Wuthrich2023,
address = {Cham},
author = {W{\"{u}}thrich, Mario V. and Merz, Michael},
doi = {10.1007/978-3-031-12409-9},
isbn = {978-3-031-12408-2},
publisher = {Springer International Publishing},
series = {Springer Actuarial},
title = {{Statistical Foundations of Actuarial Learning and its Applications}},
url = {https://link.springer.com/10.1007/978-3-031-12409-9},
year = {2023}
}

@misc{doj2022meta,
author={{U.S. Department of Justice}},
title={{United States of America v. Meta Platforms, Case 22 Civ. 5187}},
year ={2022}
}

@inproceedings{Mansoury2020FeedbackLoop,
author = {Mansoury, Masoud and Abdollahpouri, Himan and Pechenizkiy, Mykola and Mobasher, Bamshad and Burke, Robin},
title = {Feedback Loop and Bias Amplification in Recommender Systems},
year = {2020},
isbn = {9781450368599},
publisher = {Association for Computing Machinery},
address = {New York, NY, USA},
url = {https://doi.org/10.1145/3340531.3412152},
doi = {10.1145/3340531.3412152},
booktitle = {Proceedings of the 29th ACM International Conference on Information \& Knowledge Management},
pages = {2145–2148},
numpages = {4},
location = {Virtual Event, Ireland},
series = {CIKM '20}
}

@article{Mullainathan2017,
Author = {Mullainathan, Sendhil and Obermeyer, Ziad},
Title = {Does Machine Learning Automate Moral Hazard and Error?},
Journal = {American Economic Review},
Volume = {107},
Number = {5},
Year = {2017},
Month = {May},
Pages = {476–80},
DOI = {10.1257/aer.p20171084},
URL = {https://www.aeaweb.org/articles?id=10.1257/aer.p20171084}
}

@book{Muller2018Tyranny,
url = {https://doi.org/10.23943/9781400889433},
title = {The Tyranny of Metrics},
author = {Jerry Z. Muller},
publisher = {Princeton University Press},
address = {Princeton},
doi = {doi:10.23943/9781400889433},
isbn = {9781400889433},
year = {2018}
}

@article{Manheim2023Building,
  title={Building less-flawed metrics: Understanding and creating better measurement and incentive systems},
  author={Manheim, David},
  journal={Patterns},
  volume={4},
  number={10},
  year={2023},
  publisher={Elsevier}
}

@inproceedings{Raji2019Actionable,
 author = {Raji, Deborah and Denton, Emily and Bender, Emily M. and Hanna, Alex and Paullada, Amandalynne},
 booktitle = {Proceedings of the Neural Information Processing Systems Track on Datasets and Benchmarks},
 editor = {J. Vanschoren and S. Yeung},
 title = {AI and the Everything in the Whole Wide World Benchmark},
 volume = {1},
 year = {2021}
}

@inproceedings{Holstein2019improving,
author = {Holstein, Kenneth and Wortman Vaughan, Jennifer and Daum\'{e}, Hal and Dudik, Miro and Wallach, Hanna},
title = {Improving Fairness in Machine Learning Systems: What Do Industry Practitioners Need?},
year = {2019},
isbn = {9781450359702},
publisher = {Association for Computing Machinery},
address = {New York, NY, USA},
url = {https://doi.org/10.1145/3290605.3300830},
doi = {10.1145/3290605.3300830},
booktitle = {Proceedings of the 2019 CHI Conference on Human Factors in Computing Systems},
pages = {1–16},
numpages = {16},
location = {Glasgow, Scotland Uk},
series = {CHI '19}
}

@article{tomavsev2020ai,
  title={AI for social good: unlocking the opportunity for positive impact},
  author={Toma{\v{s}}ev, Nenad and Cornebise, Julien and Hutter, Frank and Mohamed, Shakir and Picciariello, Angela and Connelly, Bec and Belgrave, Danielle CM and Ezer, Daphne and Haert, Fanny Cachat van der and Mugisha, Frank and others},
  journal={Nature Communications},
  volume={11},
  number={1},
  pages={2468},
  year={2020},
  publisher={Nature Publishing Group UK London}
}

@article{floridi2021design,
  title={How to design AI for social good: Seven essential factors},
  author={Floridi, Luciano and Cowls, Josh and King, Thomas C and Taddeo, Mariarosaria},
  journal={Ethics, Governance, and Policies in Artificial Intelligence},
  pages={125--151},
  year={2021},
  publisher={Springer}
}

@article{hager2019artificial,
  author={Gregory D. Hager and Ann Drobnis and Fei Fang and Rayid Ghani and Amy Greenwald and Terah Lyons and David C. Parkes and Jason Schultz and Suchi Saria and Stephen F. Smith and Milind Tambe},
title={Artificial Intelligence for Social Good},
  journal={Computing Community Consortium Workshop Report},
  year={2019},
url={https://arxiv.org/abs/1901.05406}, 
}

@inproceedings{Ashurst2023Fairness,
author = {Ashurst, Carolyn and Weller, Adrian},
title = {Fairness Without Demographic Data: A Survey of Approaches},
year = {2023},
isbn = {9798400703812},
publisher = {Association for Computing Machinery},
address = {New York, NY, USA},
url = {https://doi.org/10.1145/3617694.3623234},
doi = {10.1145/3617694.3623234},
booktitle = {Proceedings of the 3rd ACM Conference on Equity and Access in Algorithms, Mechanisms, and Optimization},
articleno = {14},
numpages = {12},
location = {Boston, MA, USA},
series = {EAAMO '23}
}

@inproceedings{liu-etal-2023-evaluating,
    title = "Evaluating Verifiability in Generative Search Engines",
    author = "Liu, Nelson  and
      Zhang, Tianyi  and
      Liang, Percy",
    editor = "Bouamor, Houda  and
      Pino, Juan  and
      Bali, Kalika",
    booktitle = "Findings of the Association for Computational Linguistics: EMNLP 2023",
    year = "2023",
    publisher = "Association for Computational Linguistics",
    url = "https://aclanthology.org/2023.findings-emnlp.467/",
    pages = "7001--7025"
}

@article{venkit2024search,
  title={Search Engines in an AI Era: The False Promise of Factual and Verifiable Source-Cited Responses},
  author={Venkit, Pranav Narayanan and Laban, Philippe and Zhou, Yilun and Mao, Yixin and Wu, Chien-Sheng},
  journal={arXiv preprint arXiv:2410.22349},
  year={2024}
}

@misc{compound-ai-blog,
  title={The Shift from Models to Compound AI Systems},
  author={Matei Zaharia and Omar Khattab and Lingjiao Chen and Jared Quincy Davis
          and Heather Miller and Chris Potts and James Zou and Michael Carbin
          and Jonathan Frankle and Naveen Rao and Ali Ghodsi},
  howpublished={\url{https://bair.berkeley.edu/blog/2024/02/18/compound-ai-systems/}},
  year={2024}
}

@inproceedings{mirzadeh2025gsmsymbolic,
title={{GSM}-Symbolic: Understanding the Limitations of Mathematical Reasoning in Large Language Models},
author={Seyed Iman Mirzadeh and Keivan Alizadeh and Hooman Shahrokhi and Oncel Tuzel and Samy Bengio and Mehrdad Farajtabar},
booktitle={The Thirteenth International Conference on Learning Representations},
year={2025},
url={https://openreview.net/forum?id=AjXkRZIvjB}
}

@inproceedings{sainz-etal-2023-nlp,
    title = "{NLP} Evaluation in trouble: On the Need to Measure {LLM} Data Contamination for each Benchmark",
    author = "Sainz, Oscar  and
      Campos, Jon  and
      Garc{\'i}a-Ferrero, Iker  and
      Etxaniz, Julen  and
      de Lacalle, Oier Lopez  and
      Agirre, Eneko",
    editor = "Bouamor, Houda  and
      Pino, Juan  and
      Bali, Kalika",
    booktitle = "Findings of the Association for Computational Linguistics: EMNLP 2023",
    month = dec,
    year = "2023",
    address = "Singapore",
    publisher = "Association for Computational Linguistics",
    url = "https://aclanthology.org/2023.findings-emnlp.722/",
    doi = "10.18653/v1/2023.findings-emnlp.722",
    pages = "10776--10787"
}

@inproceedings{deng-etal-2024-investigating,
    title = "Investigating Data Contamination in Modern Benchmarks for Large Language Models",
    author = "Deng, Chunyuan  and
      Zhao, Yilun  and
      Tang, Xiangru  and
      Gerstein, Mark  and
      Cohan, Arman",
    editor = "Duh, Kevin  and
      Gomez, Helena  and
      Bethard, Steven",
    booktitle = "Proceedings of the 2024 Conference of the North American Chapter of the Association for Computational Linguistics: Human Language Technologies (Volume 1: Long Papers)",
    month = jun,
    year = "2024",
    address = "Mexico City, Mexico",
    publisher = "Association for Computational Linguistics",
    url = "https://aclanthology.org/2024.naacl-long.482/",
    doi = "10.18653/v1/2024.naacl-long.482",
    pages = "8706--8719"
}

@article{Shah2024Envisioning,
author = {Shah, Chirag and Bender, Emily M.},
title = {Envisioning Information Access Systems: What Makes for Good Tools and a Healthy Web?},
year = {2024},
issue_date = {August 2024},
publisher = {Association for Computing Machinery},
address = {New York, NY, USA},
volume = {18},
number = {3},
issn = {1559-1131},
url = {https://doi.org/10.1145/3649468},
doi = {10.1145/3649468},
journal = {ACM Trans. Web},
month = apr,
articleno = {33},
numpages = {24}
}

@article{lu2024ai,
  title={The ai scientist: Towards fully automated open-ended scientific discovery},
  author={Lu, Chris and Lu, Cong and Lange, Robert Tjarko and Foerster, Jakob and Clune, Jeff and Ha, David},
  journal={arXiv preprint arXiv:2408.06292},
  year={2024}
}

@article{yamada2025ai,
  title={The AI Scientist-v2: Workshop-Level Automated Scientific Discovery via Agentic Tree Search},
  author={Yamada, Yutaro and Lange, Robert Tjarko and Lu, Cong and Hu, Shengran and Lu, Chris and Foerster, Jakob and Clune, Jeff and Ha, David},
  journal={arXiv preprint arXiv:2504.08066},
  year={2025}
}

@inproceedings{Egami2023DSL,
 author = {Egami, Naoki and Hinck, Musashi and Stewart, Brandon and Wei, Hanying},
 booktitle = {Advances in Neural Information Processing Systems},
 editor = {A. Oh and T. Naumann and A. Globerson and K. Saenko and M. Hardt and S. Levine},
 pages = {68589--68601},
 publisher = {Curran Associates, Inc.},
 title = {Using Imperfect Surrogates for Downstream Inference: Design-based Supervised Learning for Social Science Applications of Large Language Models},
 url = {https://proceedings.neurips.cc/paper_files/paper/2023/file/d862f7f5445255090de13b825b880d59-Paper-Conference.pdf},
 volume = {36},
 year = {2023}
}

@article{demszky2023using,
  title={Using large language models in psychology},
  author={Demszky, Dorottya and Yang, Diyi and Yeager, David S and Bryan, Christopher J and Clapper, Margarett and Chandhok, Susannah and Eichstaedt, Johannes C and Hecht, Cameron and Jamieson, Jeremy and Johnson, Meghann and others},
  journal={Nature Reviews Psychology},
  volume={2},
  number={11},
  pages={688--701},
  year={2023},
  publisher={Nature Publishing Group US New York}
}

@inproceedings{Park2023Generative,
author = {Park, Joon Sung and O'Brien, Joseph and Cai, Carrie Jun and Morris, Meredith Ringel and Liang, Percy and Bernstein, Michael S.},
title = {Generative Agents: Interactive Simulacra of Human Behavior},
year = {2023},
isbn = {9798400701320},
publisher = {Association for Computing Machinery},
address = {New York, NY, USA},
url = {https://doi.org/10.1145/3586183.3606763},
doi = {10.1145/3586183.3606763},
booktitle = {Proceedings of the 36th Annual ACM Symposium on User Interface Software and Technology},
articleno = {2},
numpages = {22},
location = {San Francisco, CA, USA},
series = {UIST '23}
}

@article{Argyle_Busby_Fulda_Gubler_Rytting_Wingate_2023, title={Out of One, Many: Using Language Models to Simulate Human Samples}, volume={31}, DOI={10.1017/pan.2023.2}, number={3}, journal={Political Analysis}, author={Argyle, Lisa P. and Busby, Ethan C. and Fulda, Nancy and Gubler, Joshua R. and Rytting, Christopher and Wingate, David}, year={2023}, pages={337–351}}

@article{messeri2024artificial,
  title={Artificial intelligence and illusions of understanding in scientific research},
  author={Messeri, Lisa and Crockett, MJ},
  journal={Nature},
  volume={627},
  number={8002},
  pages={49--58},
  year={2024},
  publisher={Nature Publishing Group UK London}
}

@article{wang2025large,
  title={Large language models that replace human participants can harmfully misportray and flatten identity groups},
  author={Wang, Angelina and Morgenstern, Jamie and Dickerson, John P},
  journal={Nature Machine Intelligence},
  pages={1--12},
  year={2025},
  publisher={Nature Publishing Group UK London}
}

@inproceedings{Weerts2023Algorithmic,
author = {Weerts, Hilde and Xenidis, Rapha\"{e}le and Tarissan, Fabien and Olsen, Henrik Palmer and Pechenizkiy, Mykola},
title = {Algorithmic Unfairness through the Lens of EU Non-Discrimination Law: Or Why the Law is not a Decision Tree},
year = {2023},
isbn = {9798400701924},
publisher = {Association for Computing Machinery},
address = {New York, NY, USA},
url = {https://doi.org/10.1145/3593013.3594044},
doi = {10.1145/3593013.3594044},
booktitle = {Proceedings of the 2023 ACM Conference on Fairness, Accountability, and Transparency},
pages = {805–816},
numpages = {12},
location = {Chicago, IL, USA},
series = {FAccT '23}
}

@article{WACHTER2021105567,
title = {Why fairness cannot be automated: Bridging the gap between EU non-discrimination law and AI},
journal = {Computer Law \& Security Review},
volume = {41},
pages = {105567},
year = {2021},
issn = {2212-473X},
doi = {https://doi.org/10.1016/j.clsr.2021.105567},
url = {https://www.sciencedirect.com/science/article/pii/S0267364921000406},
author = {Sandra Wachter and Brent Mittelstadt and Chris Russell}
}

@article{Hellman2020Measuring,
 URL = {https://www.jstor.org/stable/27074708},
 author = {Deborah Hellman},
 journal = {Virginia Law Review},
 number = {4},
 pages = {811--866},
 publisher = {Virginia Law Review},
 title = {Measuring algorithmic fairness},
 urldate = {2025-04-29},
 volume = {106},
 year = {2020}
}

@article{Fabris2025FairnessSurvey,
author = {Fabris, Alessandro and Baranowska, Nina and Dennis, Matthew J. and Graus, David and Hacker, Philipp and Saldivar, Jorge and Zuiderveen Borgesius, Frederik and Biega, Asia J.},
title = {Fairness and Bias in Algorithmic Hiring: A Multidisciplinary Survey},
year = {2025},
issue_date = {February 2025},
publisher = {Association for Computing Machinery},
address = {New York, NY, USA},
volume = {16},
number = {1},
issn = {2157-6904},
url = {https://doi.org/10.1145/3696457},
doi = {10.1145/3696457},
journal = {ACM Trans. Intell. Syst. Technol.},
month = jan,
articleno = {16},
numpages = {54}
}

@article{Loi_Herlitz_Heidari_2024,
title={Fair equality of chances for prediction-based decisions}, volume={40}, DOI={10.1017/S0266267123000342}, number={3}, journal={Economics and Philosophy}, author={Loi, Michele and Herlitz, Anders and Heidari, Hoda}, year={2024}, pages={557–580}}

@inproceedings{Alvarez2023Counterfactual,
author = {Alvarez, Jose Manuel and Ruggieri, Salvatore},
title = {Counterfactual Situation Testing: Uncovering Discrimination under Fairness given the Difference},
year = {2023},
isbn = {9798400703812},
publisher = {Association for Computing Machinery},
address = {New York, NY, USA},
url = {https://doi.org/10.1145/3617694.3623222},
doi = {10.1145/3617694.3623222},
booktitle = {Proceedings of the 3rd ACM Conference on Equity and Access in Algorithms, Mechanisms, and Optimization},
articleno = {2},
numpages = {11},
location = {Boston, MA, USA},
series = {EAAMO '23}
}

@inproceedings{NIPS2017_3f5ee243,
 author = {Vaswani, Ashish and Shazeer, Noam and Parmar, Niki and Uszkoreit, Jakob and Jones, Llion and Gomez, Aidan N and Kaiser, \L ukasz and Polosukhin, Illia},
 booktitle = {Advances in Neural Information Processing Systems},
 editor = {I. Guyon and U. Von Luxburg and S. Bengio and H. Wallach and R. Fergus and S. Vishwanathan and R. Garnett},
 pages = {},
 publisher = {Curran Associates, Inc.},
 title = {Attention is All you Need},
 url = {https://proceedings.neurips.cc/paper_files/paper/2017/file/3f5ee243547dee91fbd053c1c4a845aa-Paper.pdf},
 volume = {30},
 year = {2017}
}

\clearpage

\chapter{{\color{goalIcolor}Measure: Fairness Metrics and Frameworks}}
\label{chaptergoal1}

\section{Fairness in Algorithmic Insurance Pricing}
\label{ssec:paper1}

\vspace{10mm}
This paper is published as:
\\
\begin{myquote}
\paperIcitationnew
\end{myquote}

\includepdf[pages=-]{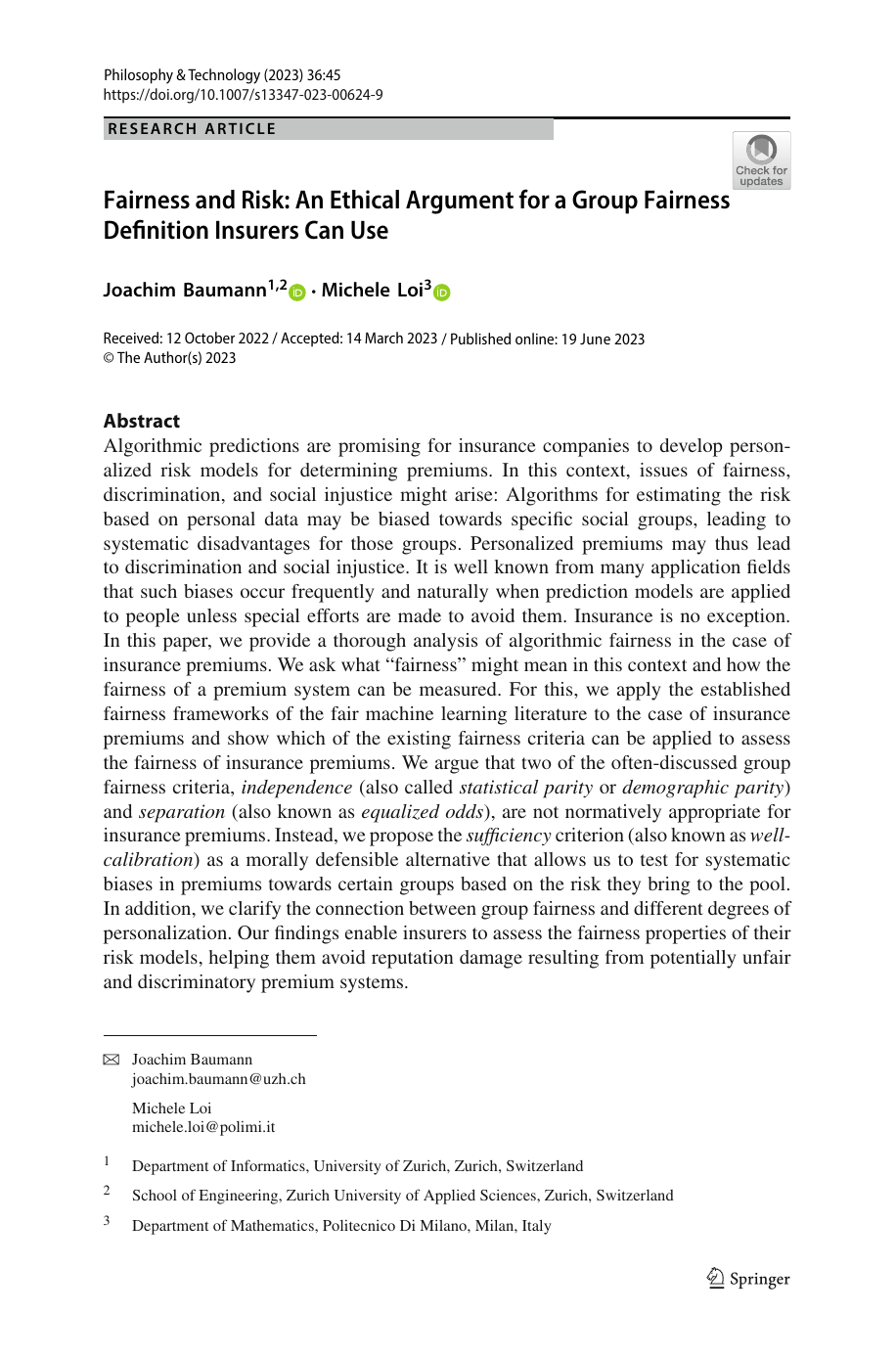}

\clearpage

\section{A Comprehensive Framework for Group Fairness Metrics}
\label{ssec:paper2}

\vspace{10mm}
A short version of this paper is published as:
\\
\begin{myquote}
\paperIIcitationnew
\end{myquote}
\vspace{7mm}
\noindent
Here we include an extended version of this paper, which was presented as a poster at the \textit{3rd ACM Conference on Equity and Access in Algorithms, Mechanisms, and Optimization (EAAMO'2023) (non-archival)}~\citep{baumann2022distributive}.

\includepdf[pages=-]{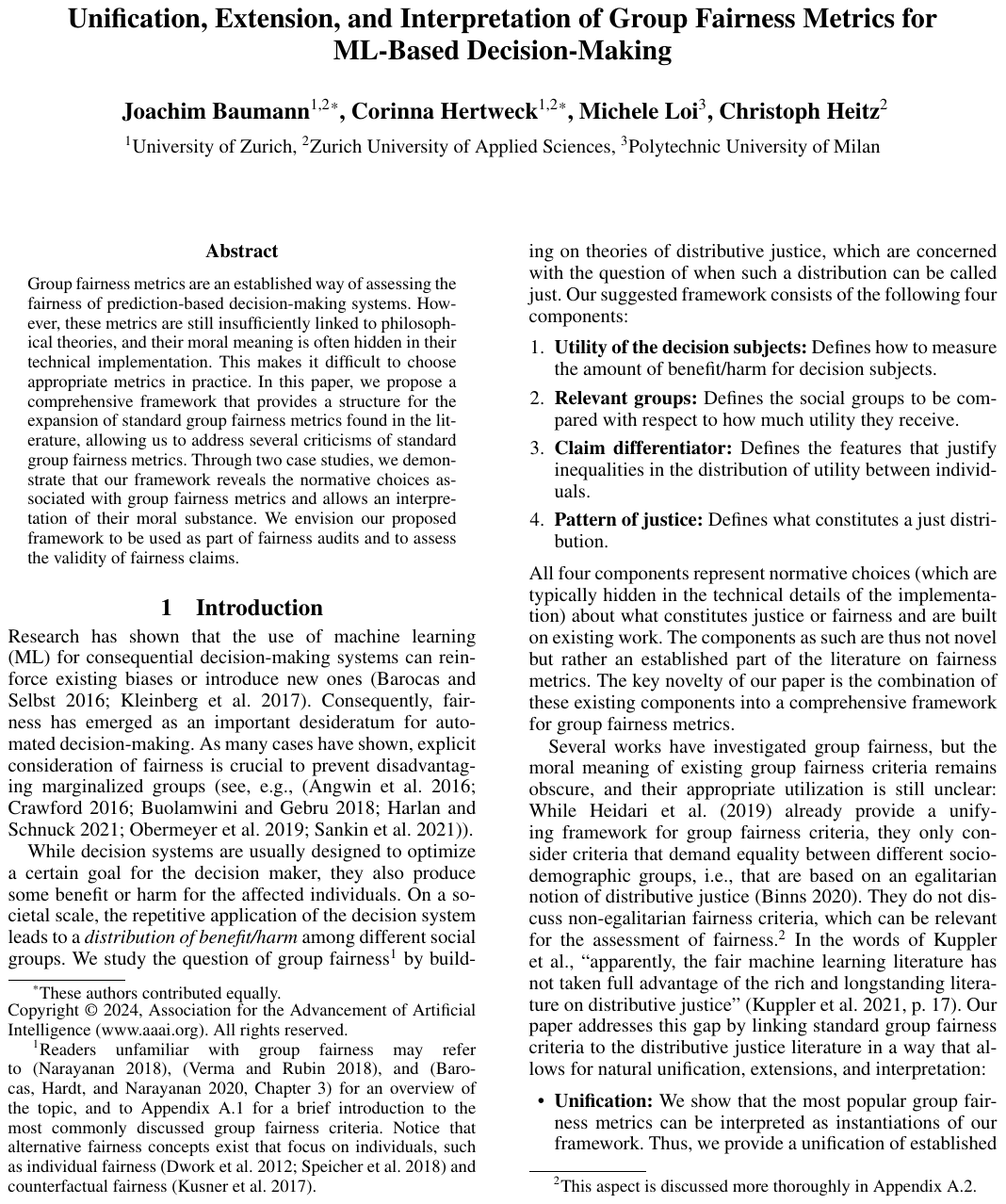}

\clearpage

\chapter{{\color{goalIIcolor}Decompose: Biases and Feedback Loops}}
\label{chaptergoal2}

\section{Synthetic Data Generation for Bias Analysis}
\label{ssec:paper3}

\vspace{10mm}
This paper is published as:
\\
\begin{myquote}
\paperIIIcitationnew
\end{myquote}

\includepdf[pages=-]{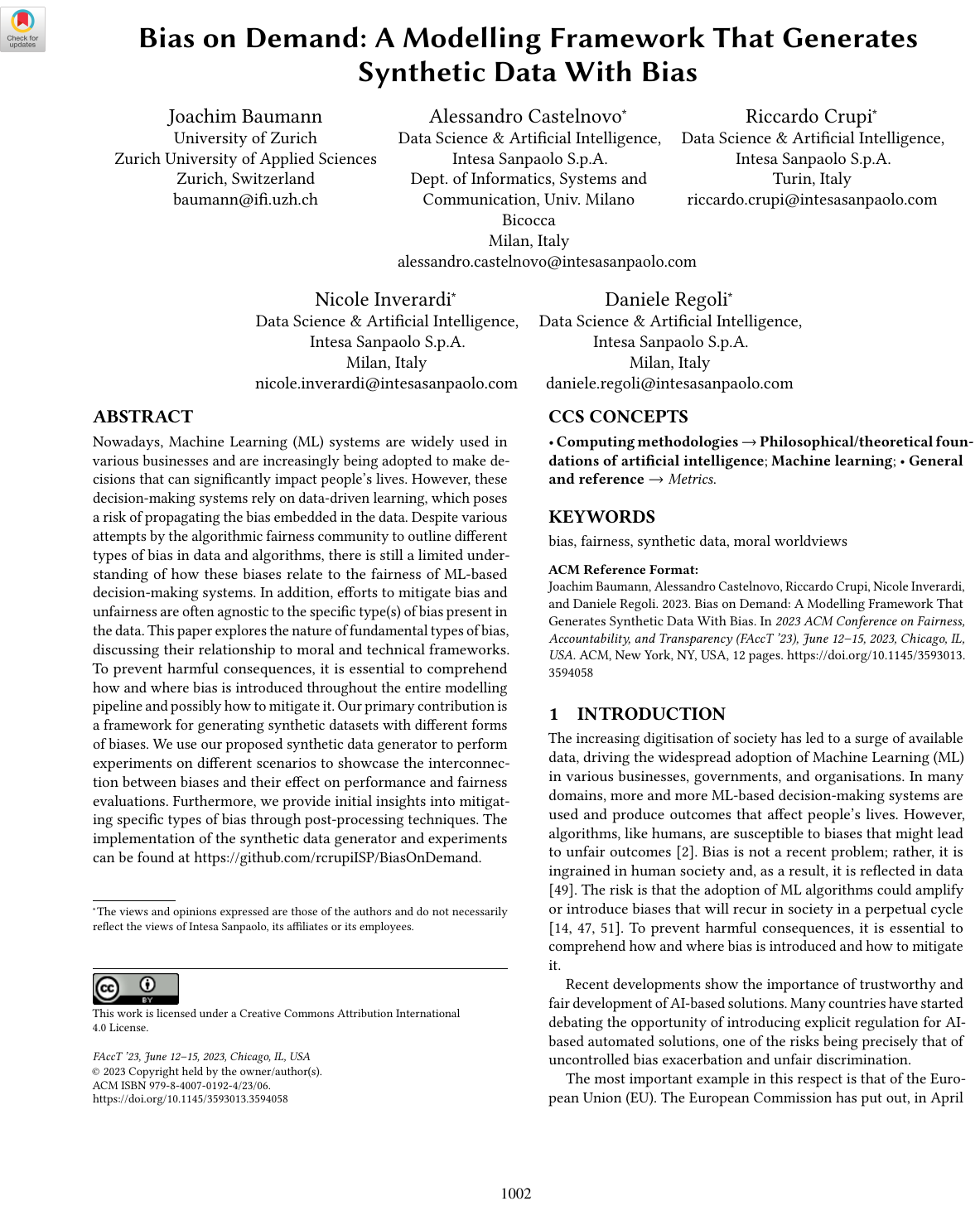}
\includepdf[pages=-]{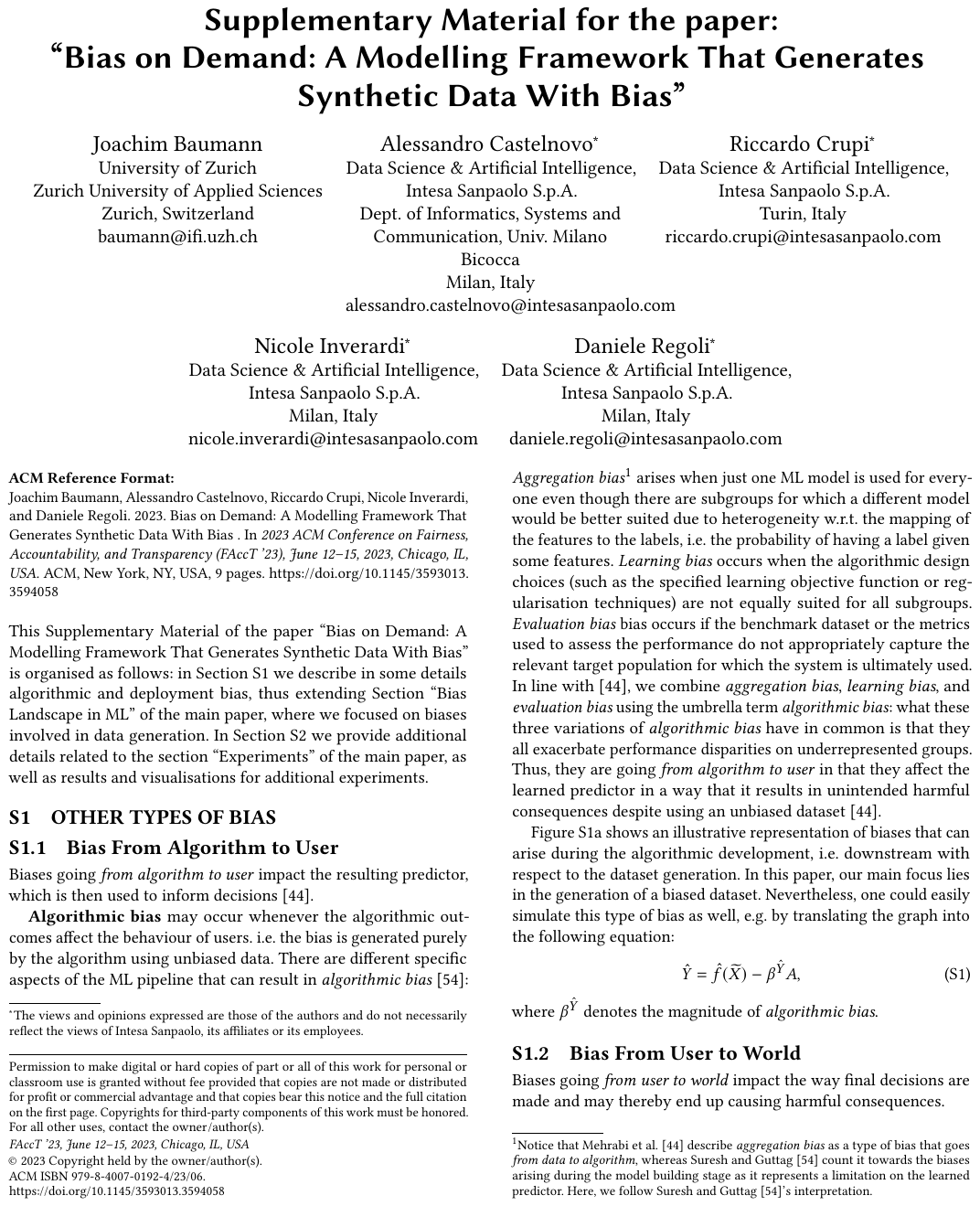}

\clearpage

\section{Classification of Feedback Loops in ML Systems}
\label{ssec:paper4}

\vspace{10mm}
This paper is published as:
\\
\begin{myquote}
\paperIVcitationnew
\end{myquote}

\includepdf[pages=-]{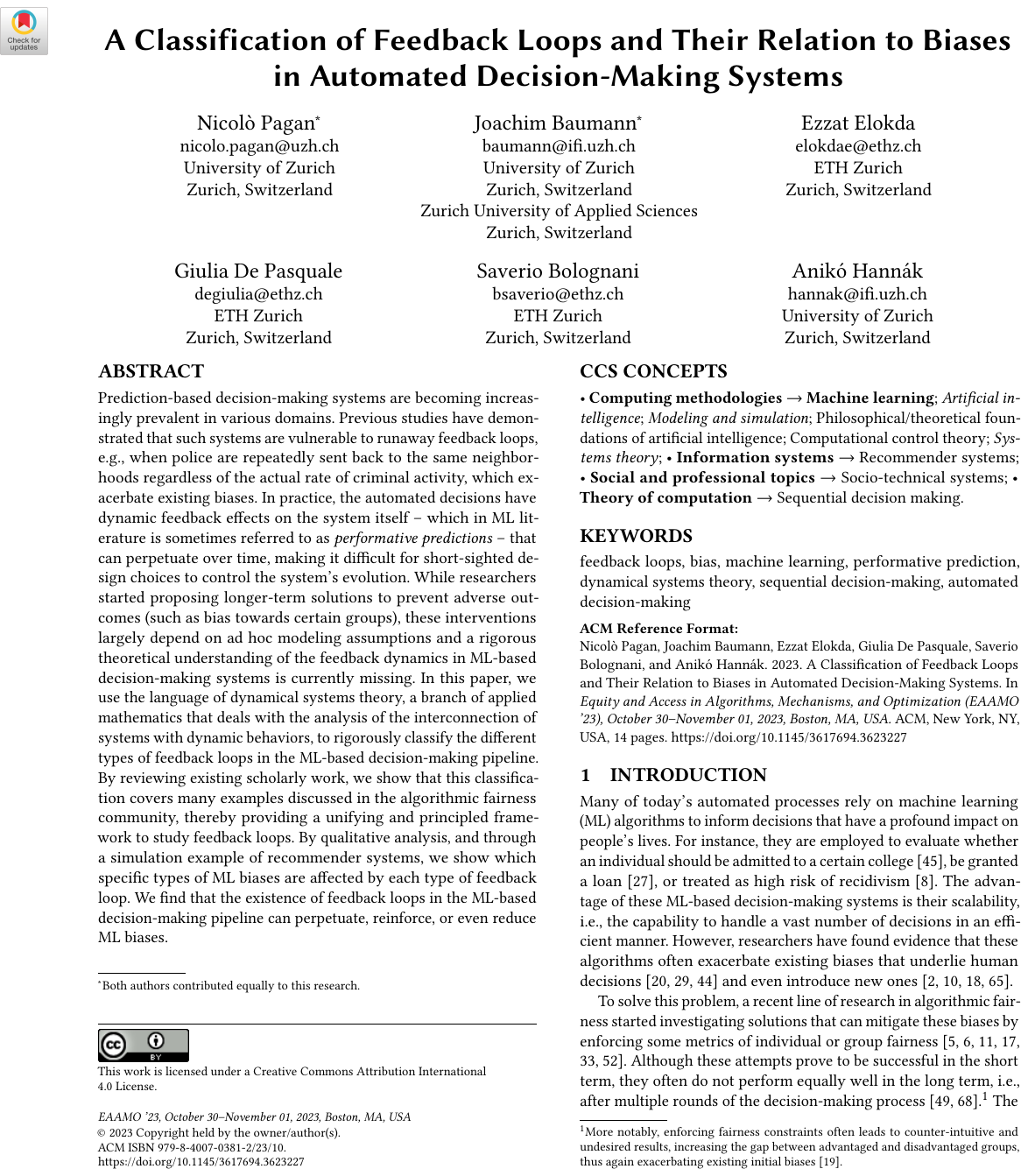}

\clearpage

\section{Fairness Trade-offs in Online Advertising}
\label{ssec:paper5}

\vspace{10mm}
This paper is published as:
\\
\begin{myquote}
\paperVcitationnew
\end{myquote}

\includepdf[pages=-]{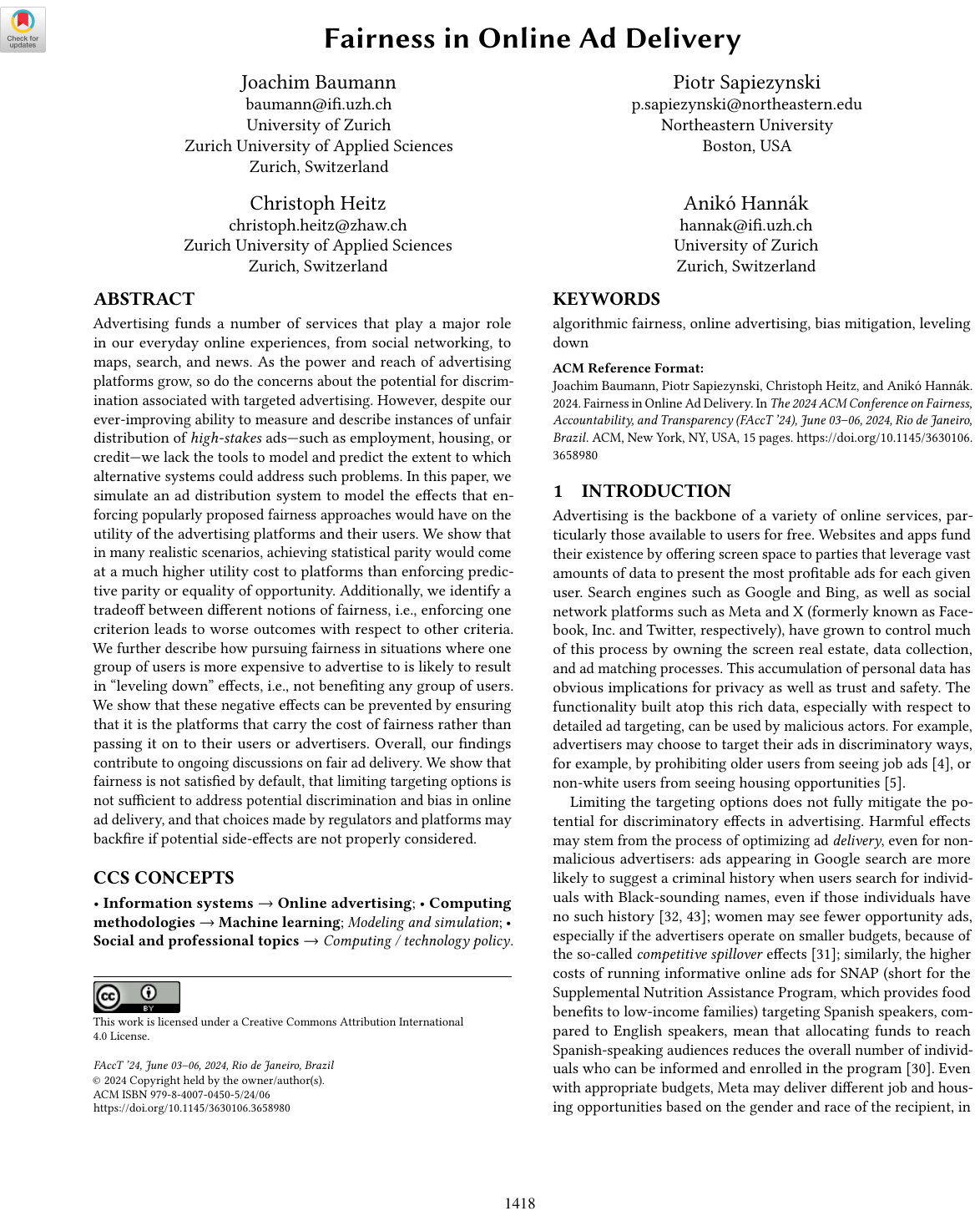}

\clearpage

\chapter{{\color{goalIIIcolor}Fix: Mitigation Strategies and Applications}}
\label{chaptergoal3}

\section{Optimal Decision Making Under Fairness Constraints}
\label{ssec:paper6}

\vspace{10mm}
This paper is published as:
\\
\begin{myquote}
\paperVIcitationnew
\end{myquote}

\includepdf[pages=-]{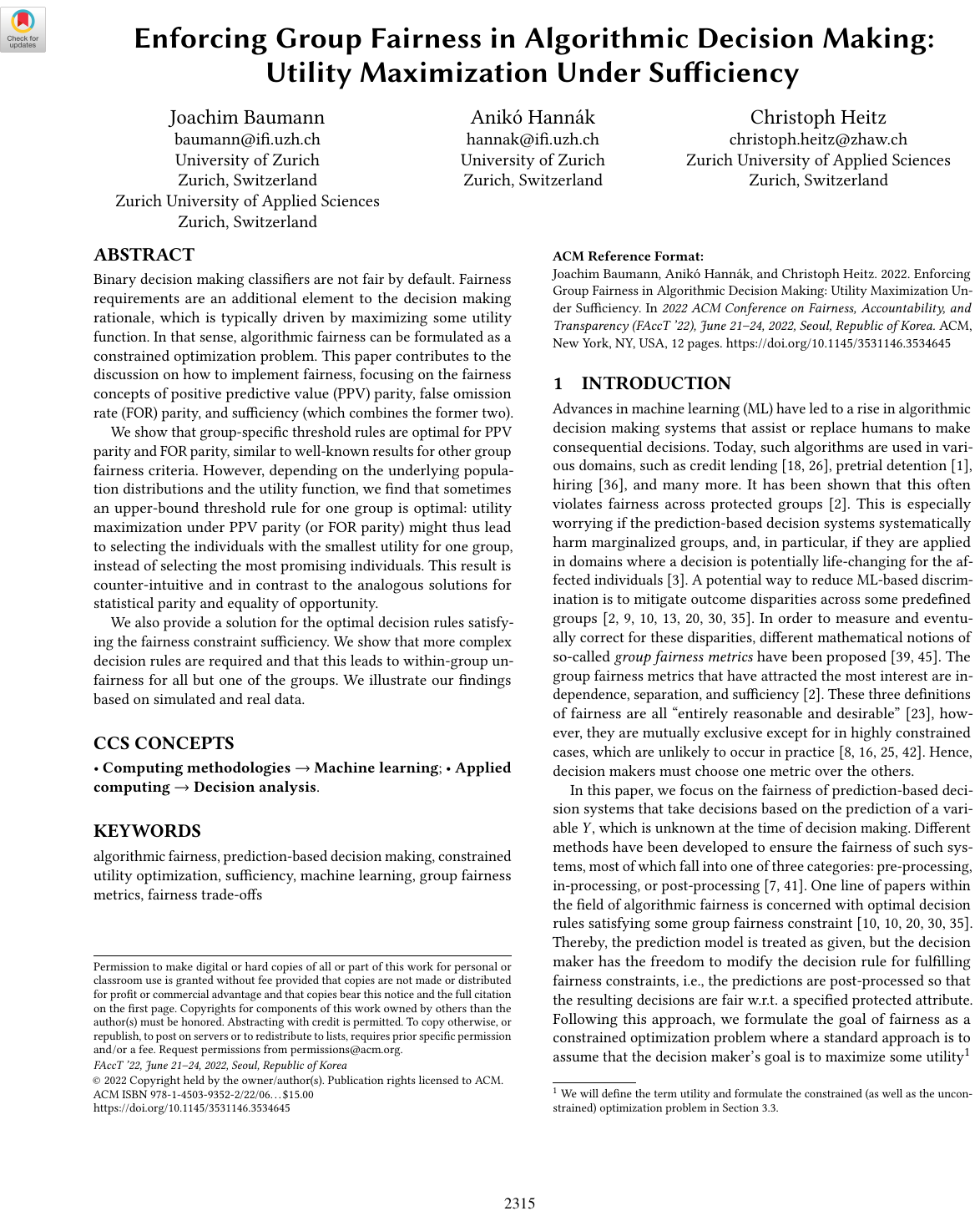}
\includepdf[pages=-]{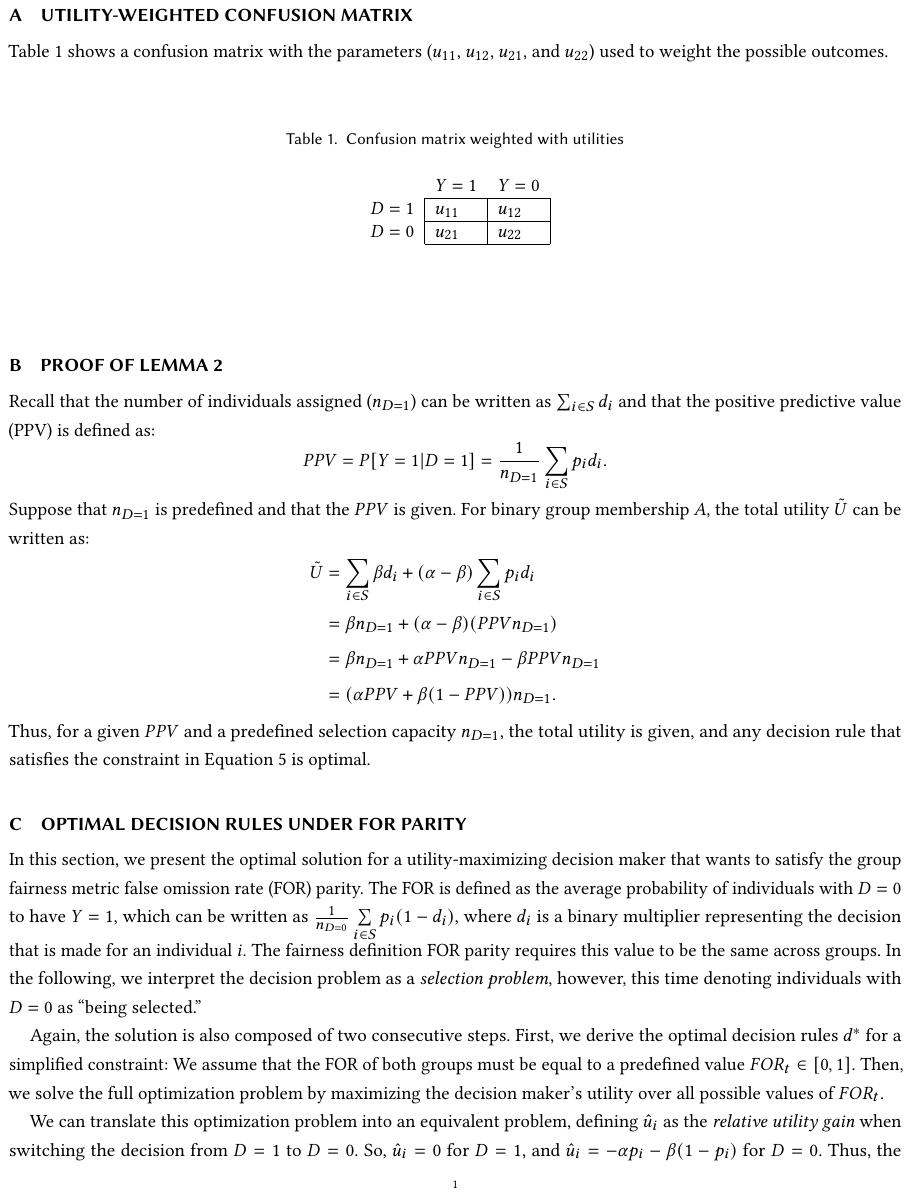}

\clearpage

\section{Algorithmic Collective Action in Recommender Systems}
\label{ssec:paper7}

\vspace{10mm}
This paper is published as:
\\
\begin{myquote}
\paperVIIcitationnew
\end{myquote}

\includepdf[pages=-]{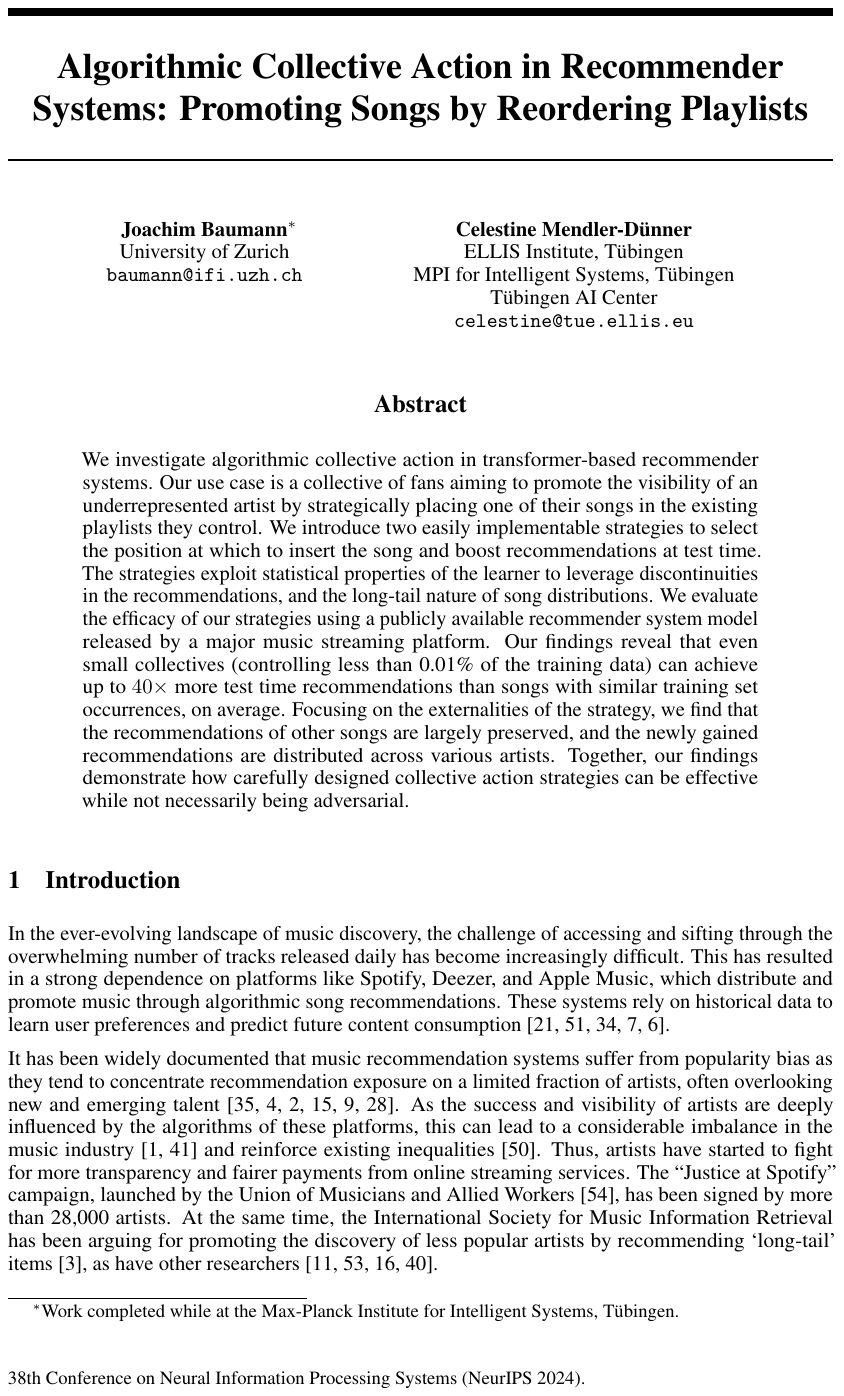}

\clearpage

\section{ML for Equitable Rental Assistance Allocation}
\label{ssec:paper8}

\vspace{10mm}
This paper is published as:
\\
\begin{myquote}
\paperVIIIcitationnew
\end{myquote}

\includepdf[pages=-]{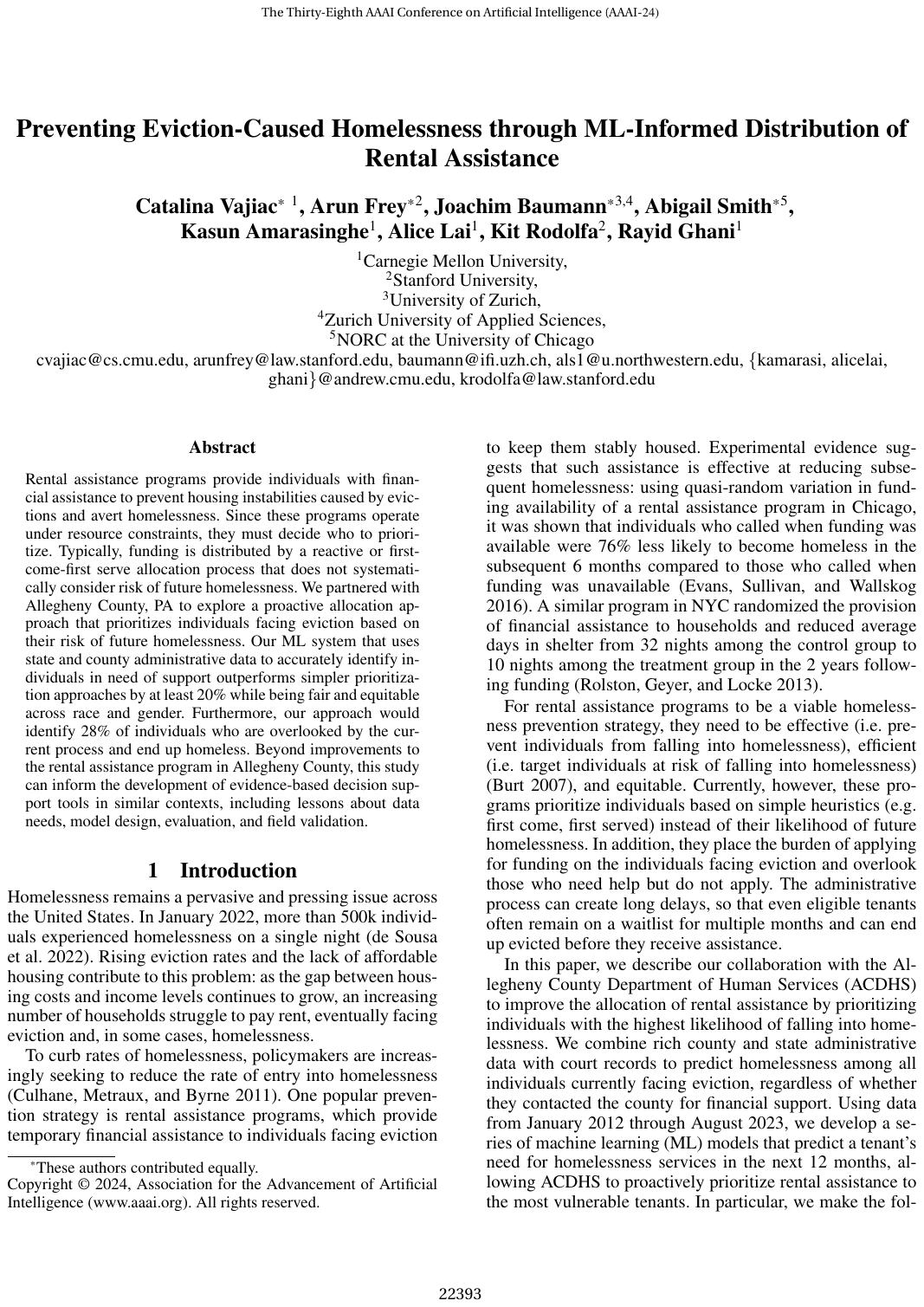}
\includepdf[pages=-]{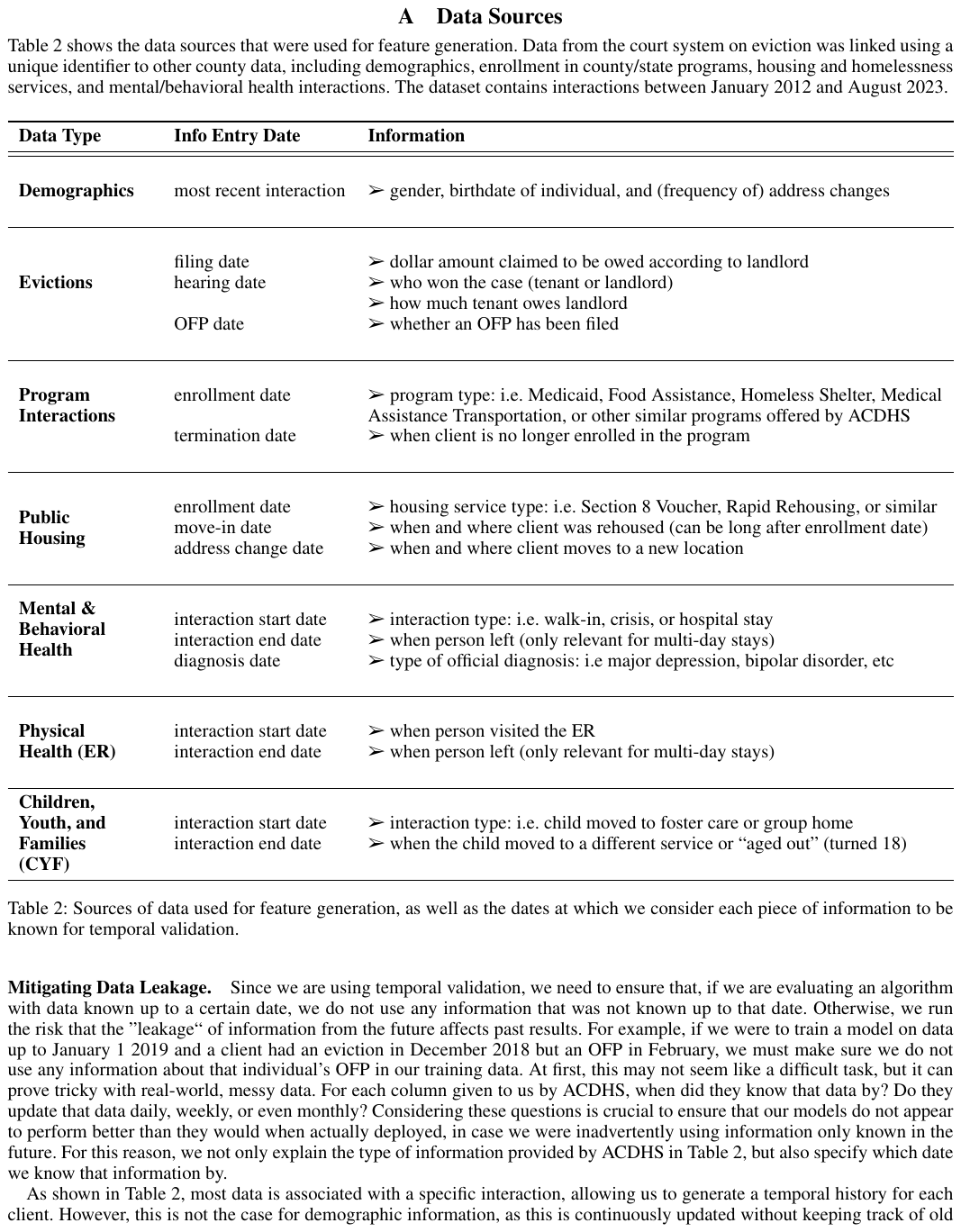}

\clearpage

\section*{Curriculum Vitae}

\subsection*{Education}
\begin{tabular}{ll}
           & \textbf{University of Zurich}\\
2021--2025 & PhD Student in Informatics\\
2021--2023 & Excellence Program for PhD Students -- Digital Society Initiative\\
2019--2021 & Master of Science in Informatics \textit{(summa cum laude)}\\
2015--2019 & Bachelor of Science in Informatics \textit{(magna cum laude)}\\
           & \textbf{Universidad de Guadalajara}\\
2020        & Exchange Semester\\
           & \textbf{Universitat Pompeu Fabra Barcelona}\\
2017--2018 & Erasmus Exchange Semester\\
\end{tabular}

\subsection*{Professional Experience}
\begin{tabular}{ll}
           & \textbf{Bocconi University}\\
2025        & Postdoc -- Milan Natural Language Processing Lab (MilaNLP)\\
           & \textbf{Max Planck Institute for Intelligent Systems in Tübingen}\\
2023--2024 & PhD Research Intern -- Social Foundations of Computation Department\\
           & \textbf{University of Zurich}\\
2019--2025 & Research Assistant (non-consecutive)\\
           & \textbf{Zurich University of Applied Sciences}\\
2019--2024 & Research Assistant\\
           & \textbf{Carnegie Mellon University}\\
2022        & Data Science for Social Good Fellow\\
           & \textbf{Campana \& Schott Switzerland AG}\\
2018--2019 & Technology Consulting\\
           & \textbf{Lifeware SA Zurich}\\
2014--2015 & Application Development\\
\end{tabular}

\end{document}